%% file: main.tex
\documentclass[journal]{IEEEtran}
\usepackage{amsmath,amsfonts}
\usepackage{algorithmic}
\usepackage{algorithm}
\usepackage{array}
\usepackage[caption=false,font=normalsize,labelfont=sf,textfont=sf]{subfig}
\usepackage{textcomp}
\usepackage{stfloats}
\usepackage{url}
\usepackage{verbatim}
\usepackage{graphicx}
\usepackage{cite}
\usepackage{xcolor}
\usepackage{xspace}
\usepackage{hyperref}
\def\ours{AllDayNav\xspace}

\begin{document}

\title{\ours: Lifelong Navigation via Real-World Reinforcement Learning}

\author{Hang Yin*, Yinan Liang*, Jiazhao Zhang*, Jiahang Liu, Minghan Li, Zhizheng Zhang$\dagger$, and He Wang$\dagger$\thanks{
Hang Yin and Yinan Liang are with Tsinghua University, Beijing 100084, China, and also with Galbot Robotics, Beijing, China (e-mail: yinh23@mails.tsinghua.edu.cn; liangyn24@mails.tsinghua.edu.cn).~
Jiazhao Zhang is with Peking University, Beijing 100871, China, and also with Galbot Robotics, Beijing, China (e-mail: zhngjizh@gmail.com).~
Jiahang Liu and Minghan Li are with Galbot Robotics, Beijing, China (e-mail: liu030526@gmail.com;lperlpm@gmail.com).~
Zhizheng Zhang is with Galbot Robotics, Beijing, China, and also with Beijing Academy of Artificial Intelligence, Beijing, China (e-mail: zhizheng@mail.ustc.edu.cn).~
He Wang is with Galbot Robotics, Beijing, China, Peking University, Beijing 100871, China, and also with Beijing Academy of Artificial Intelligence, Beijing, China (e-mail: hewang@pku.edu.cn).~
*Equal contribution $\dagger$ Corresponding authors
}}



\IEEEaftertitletext{\vspace{-1\baselineskip}
\begin{center}
        \includegraphics[width=0.95\textwidth]{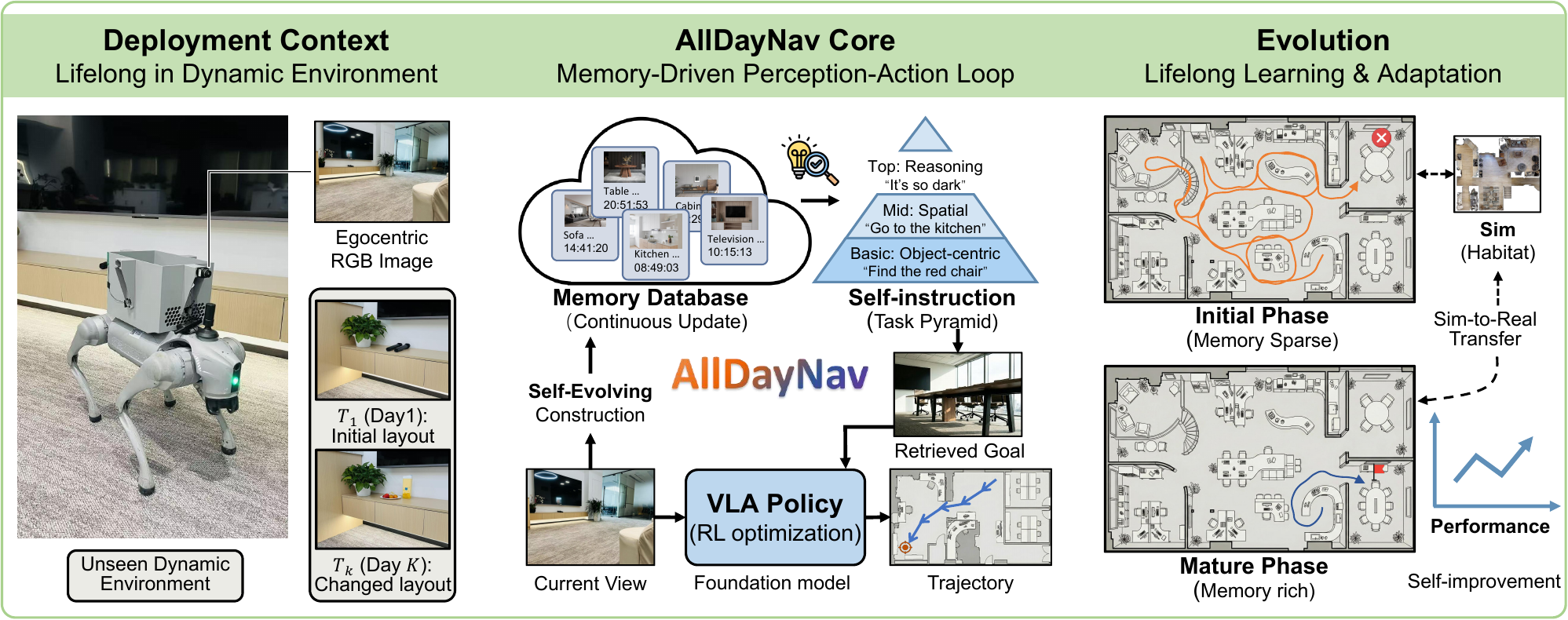}
        \vspace{0.3cm}
        
        \parbox{0.95\textwidth}{\small Fig. 1. Overview of \textbf{\ours}, a lifelong self-learning navigation framework. \textbf{(Left)} Deployment in unseen dynamic environments (simulated and real-world). \textbf{(Middle)} The core \textbf{Memory--Policy Co-evolution}: \ours autonomously builds a self-evolving multimodal memory database to generate self-instructions and retrieve visual goals, training a Vision-Language-Action (VLA) policy without supervision. \textbf{(Right)} As experience accumulates, the robot masters lifelong navigation and adapts to environmental changes.}
        \label{fig:teaser}
\end{center}
\vspace{1em}
\stepcounter{figure}
}

\maketitle

\input{sections/0_abstract}

\input{sections/1_intro}
\input{sections/2_related_works}
\input{sections/3_problem_statement}
\input{sections/4_method}

\input{sections/5_experiment}
\input{sections/6_conclusion}


{
    \bibliographystyle{IEEEtran}
    \bibliography{main}
}

\newpage

\vfill

\end{document}

%% file: sections/0_abstract.tex
\begin{abstract}
Lifelong embodied navigation in dynamic environments requires robots to form persistent scene understanding from fragmentary observations, which remains difficult for existing methods that rely on explicit maps or scene graphs and struggle to generalize beyond structured settings.
We propose \ours, a lifelong self-learning navigation framework that implicitly encodes scene dynamics into the billion-scale parameters of a large model via reinforcement learning, powered by a self-evolving multimodal memory that maintains and updates visual keyframes, semantic descriptions, and temporal context while autonomously generating open-vocabulary instructions, image goals, and structured rewards.
Experiments in both synthetic and real-world environments across cross-room, cross-episode, and cross-task scenarios show that \ours achieves success rates approaching $100\%$ and consistently surpasses strong map-based, VLM, and RL baselines in path efficiency and robustness, demonstrating implicit, memory-driven reinforcement learning as a scalable alternative to explicit mapping for reliable lifelong navigation. \href{https://bagh2178.github.io/AllDayNav/}{\textcolor[rgb]{1,0.4,0.6}{Project Page.}}
\end{abstract}


%% file: sections/1_intro.tex
\section{Introduction}
\IEEEPARstart{L}{ifelong} navigation in real-world environments requires embodied agents to memorize and continually update their understanding of dynamic surrounding environments, as they must operate across extended periods without assuming static scene structure. Such settings demand that an agent maintain a temporally coherent internal representation despite ambiguous observations, environmental changes, and limited field of view \cite{wu2024embodied, anderson2018evaluation}. To support these requirements, recent vision–language–action (VLA) foundation models \cite{anderson2018vision, qiao2024llm, zhang2024navid} provide strong perceptual and semantic reasoning capabilities, enabling open-vocabulary goal interpretation \cite{guadarrama2016understanding, guadarrama2014open, yokoyama2024hm3d}, scene understanding \cite{fan2022vision, zhang2022outdoor, humblot2022navigation}, and instruction-conditioned action generation \cite{li2025cogvla, kim2025multi, lee2025dynscene}. Leveraging large-scale multimodal pretraining \cite{ramakrishnan2021habitat, chang2017matterport3d, xia2018gibson}, these models have demonstrated strong zero-shot generalization across diverse navigation tasks \cite{chaplot2020object, pal2021learning}, suggesting that they offer a viable foundation for addressing the challenges posed by lifelong navigation.

Despite the need for persistent and adaptable memory, current VLA-based navigation systems still operate with short frame-based memory \cite{liu2025nvila, qiao2025open}, finetuned scene-specific representations \cite{hong2025general, kuang2024openfmnav}, or static scene graphs \cite{yin2025unigoal, singh2023scene}. These memories remain narrowly scoped and static, which prevents them from providing the persistent and adaptable memory required for lifelong navigation, especially in dynamic environments. Beyond memory-based approaches, current VLA adaptation strategies rely primarily on episodic RL finetuning \cite{lu2025vla, li2025vla} or limited-interaction online updates \cite{li2025simplevla, guo2025improving} . These adaptations remain local and episode-bound, which prevents policies from accumulating knowledge or maintaining globally consistent behavior over extended deployments. These limitations highlight the need for a representational principle that naturally supports persistent, visually grounded, and lifelong coherent navigation.

Cognitive neuroscience indicates~\cite{kosslyn2003mental, kosslyn1978visual} that human navigation is supported by visual mental imagery that provides a stable internal depiction of the environment and enables both spatial recall and action planning~\cite{borst2010individual, farah1988visual}. Mental imagery is reconstructive and temporally stable, enabling people to update and refine their internal representation of the environment, which also provides an evolving visual memory \cite{keogh2011mental}. This perspective suggests that effective embodied agents should maintain a similarly imagery-like internal representation that both preserves visual continuity and informs policy refinement throughout lifelong navigation. 

Motivated by this perspective, we introduce \ours, a lifelong self-learning embodied navigation framework that implements imagery-inspired internal representations for embodied agents. \ours represents navigation goals as visual targets rather than symbolic instructions, and it constructs a persistent internal memory consistent with this imagery-based representation. As the agent explores, \ours incrementally builds a multimodal memory that stores imagery-aligned visual embeddings together with semantic descriptions, categorical information, and temporal context. This memory is refined continuously during interaction, enabling the agent to retain coherent visual and semantic information across episodes and environmental changes.

To couple memory with policy learning, \ours employs a memory-driven reinforcement learning framework that integrates perception, memory updating, retrieval, and action generation into a unified decision-making process. This integration allows the agent to refine its behavior using its evolving internal representation, analogous to how humans use visual imagination to anticipate future states and plan movement. Collectively, these components enable \ours to convert fragmentary egocentric observations into a stable internal representation that supports reliable lifelong navigation in complex, real-world environments. We evaluate \ours extensively in both simulated and real-world environments across diverse lifelong embodied navigation tasks. Experiments show consistent and substantial gains over VLM/LLM-based navigation systems, map-centric memory approaches, and reinforcement-learning baselines in accuracy, robustness, and generalization, with success rates approaching 100\% in unseen environments. \ours maintains reliable performance in unknown and dynamically changing environments through its imagery-inspired memory and adaptive policy learning. Our main contributions are as follows:

\begin{itemize}
    \item We introduce \ours, a navigation framework motivated by visual mental imagery that represents goals in an image-centric form and maintains a coherent internal depiction of the environment for lifelong self-learning embodied navigation.

    \item We develop a persistent multimodal memory architecture that records imagery-oriented visual embeddings, semantic attributes, and temporal context across episodes, providing a structured and continually evolving representation of past experience.

    \item We design a memory-driven reinforcement learning framework that integrates the learned memory into the control loop, enabling policies to adapt through interaction and improve autonomously without human supervision.
\end{itemize}

%% file: sections/2_related_works.tex
\section{Related Work}
\subsection{Memory-Based Navigation}
Lifelong navigation depends on memory to fuse partial observations into a coherent environmental representation. Early approaches adopted structured, explicit scene representations \cite{liu2023bird, ravichandran2022hierarchical, chi2025navigating}, in which the environment is encoded through semantic maps or hierarchical scene graphs. Systems such as 3D Scene Graph \cite{strader2024indoor} organize objects, rooms, and spatial relations into multi-layer graph structures, providing an interpretable and spatially consistent representation for downstream reasoning. Extensions such as Open Scene Object Navigation \cite{loo2025open} leverage open-vocabulary vision-language models to construct scene graphs that remain applicable in previously unseen environments. Methods such as SG-Nav \cite{yin2024sg} couple hierarchical 3D scene graph construction with LLM-based chain-of-thought reasoning, enabling the selection of informative navigation frontiers during exploration. Although these structured memories provide clear interpretability and explicit spatial grounding, their updates are computationally intensive, limiting scalability in dynamic or lifelong deployments.

To address the scalability constraints of explicit maps, recent work has explored embedding-based \cite{majumdar2022zson, kotar2023entl} and retrieval-augmented \cite{monaci2025rana, glocker2025llm} memory systems that encode visual, linguistic, and spatial observations into compact vector representations. For instance, ReMEmbR \cite{anwar2025remembr} introduces a retrieval-augmented spatio-temporal memory in which each observation is stored as a multimodal embedding and later recalled to inform decision-making. Meta-Memory \cite{mao2025meta} extends this paradigm by using dual-channel retrieval to jointly leverage semantic and spatial cues, enabling the agent to integrate new observations with existing knowledge into a continually updated cognitive map. Compared with structured memories, retrieval-based systems offer better scalability, operate consistently across modalities, and are more practical for lifelong deployment.

However, a common limitation across both structured and embedding-based memories is that memory construction remains decoupled from policy learning. Memories are typically built through passive accumulation or fixed pretrained encoders and thus remain static during downstream interaction. Consequently, agents cannot modify either the content of their memories or the retrieval strategies they use in response to task rewards, exploration dynamics, or changing objectives. This disconnect prevents memory representations from being refined through continual experience and limits the development of task-specific memory structures needed for lifelong improvement.

\subsection{Large Language Models for Navigation}
Large language models \cite{chiang2023vicuna, yang2025qwen3} have been introduced into embodied navigation to provide semantic interpretation of natural-language instructions and high-level task reasoning. Early LLM-based navigation systems \cite{zhu2024chatnav, chen2024mapgpt, long2024instructnav} rely on symbolic, text-centered pipelines in which the model decomposes natural-language instructions into low-level actions. Systems such as NavGPT \cite{zhou2024navgpt} implement an iterative dialogue between a pretrained LLM and an environment parser, whereas NavCoT \cite{lin2025navcot} augments this pipeline with chain-of-thought prompts that explicitly articulate intermediate semantic steps. Although these approaches provide interpretable decision processes, their reliance on text-only abstractions creates a disconnect between visual perception and action generation and constrains their ability to operate on continuous visual input.

To reduce this dependence on symbolic intermediates, subsequent work has shifted toward video-based multimodal vision-language-action models \cite{xu2024mobility, yokoyama2025film} that jointly encode egocentric visual streams and textual inputs within a unified representation. This design provides tighter perception-action coupling and enables temporally coherent reasoning. Uni-NaVid \cite{zhang2024uni} demonstrates that a shared multimodal model can support a wide range of visuomotor tasks by processing video and language tokens within a common embedding space. Navigation World Model \cite{bar2025navigation} formulates navigation as a video prediction problem, using a generative transformer to model environment dynamics and forecast future trajectories directly from raw visual observations. These models substantially improve spatial grounding and sequential prediction capability relative to purely text-conditioned planners.

Despite these advances, LLM planners and multimodal VLA models remain stateless. Each episode is processed independently, and information obtained during interaction is discarded afterward. Without persistent multimodal memory, these systems cannot accumulate experience, preserve contextual continuity across episodes, or adapt their behavior based on past interactions. This lack of persistent state limits their capacity for continual improvement and long-term autonomous operation in embodied environments.

\subsection{Reinforcement Learning for Embodied Agents}
Reinforcement learning enables interaction-driven policy adaptation, yet most recent work uses it primarily to fine-tune pretrained VLA models \cite{chen2025conrft, luo2024serl} rather than to support continual learning. Methods such as VLA-RL \cite{lu2025vla} apply trajectory-level policy optimization on top of pretrained multimodal encoders to improve task performance while maintaining semantic grounding. SimpleVLA-RL \cite{li2025simplevla} replaces imitation objectives with RL losses to enhance robustness and compositional generalization in manipulation tasks. Related approaches, including VLA-RFT \cite{li2025vla} and reward-aligned instruction finetuning, use structured reward signals or human feedback to guide policy updates so that agent behavior remains consistent with high-level task semantics. Although these techniques improve the performance of VLA-based agents, their learning remains constrained to fixed datasets or limited interaction logs, preventing policies from retaining experience across episodes and limiting adaptation to lifelong navigation problems.

In a complementary direction, another line of work examines continual and real-world RL, where policies are updated from sustained interaction with the physical environment. Classical systems such as Real-World RL \cite{zhu2020ingredientsrealworldroboticreinforcement} and SERL \cite{luo2024serl} introduce safety-aware pipelines that enable sample-efficient online learning through automatic resets and off-policy updates. Building on these principles, more recent frameworks such as RLIF \cite{luo2023rlif}, RLDG \cite{xu2024rldg}, and PoliFormer \cite{zeng2024poliformer} combine RL with foundation models or world modeling to enable knowledge transfer across tasks and domains. These systems highlight the potential of continual RL for long-term adaptation in dynamic settings.

Despite these advances, RL-based embodied agents remain limited by the lack of high-level semantic structure and persistent multimodal memory. Fine-tuned VLA models rely on strong perception and language priors but do not retain knowledge across episodes, while continual RL systems improve through interaction but lack mechanisms for structured reasoning or for storing and retrieving past experience. These limitations reduce an agent's ability to build stable long-term knowledge and to coordinate semantic reasoning with adaptive behavior in complex environments.

%% file: sections/3_problem_statement.tex
\section{Problem Formulation}

This section provides a formal definition of the lifelong self-learning embodied navigation task in open-world indoor environments.
We begin by describing the task overview, followed by a mathematical formulation that decomposes the problem into pretraining and online self-learning phases, and conclude with an analysis of the core technical constraints and challenges. 

\subsection{Task Overview}

Lifelong navigation considers an embodied agent operating in previously unseen indoor environments over extended temporal horizons. The agent must form and maintain an internal representation that remains coherent across episodes while accommodating structural and semantic changes that may arise during deployment. This setting departs from conventional episodic navigation, which assumes short horizons, static scenes, and full state resets, and instead requires persistent understanding and continual adaptation.

We study an agent deployed in an initially unknown and potentially dynamic indoor environment \(E\). At each timestep \(t\), the agent receives an egocentric RGB observation \(o_t \in \mathbb{R}^{H \times W \times 3}\) together with an open-vocabulary natural-language instruction \(I\). The objective is to learn a navigation policy \(\pi\) that outputs continuous waypoints or locomotion actions guiding the agent to a location that satisfies the semantics of \(I\), without relying on predefined goal categories, prior maps, demonstrations, or reward annotations. 

A defining characteristic of this setting is that the agent must construct and maintain a persistent understanding of the scene from fragmented egocentric observations while operating over long, cross-episode horizons. In addition, the agent must interpret open-vocabulary instructions by grounding them into visual targets and continually refine its behavior through self-instructed experience rather than externally curated supervision. Task success is recorded when the agent autonomously stops at a viewpoint that semantically corresponds to the instructed goal under lifelong operation.

\subsection{Lifelong Learning Overview}

We implement the lifelong self-learning embodied navigation problem as two phases: an offline \emph{pretraining phase} in which the agent learns foundational navigation skills on large-scale datasets, and an online \emph{self-learning phase} in which the agent refines its policy through autonomous interaction with the target environment. We elaborate on these two phases as follows:

\subsubsection{Pretraining Phase}

In the pretraining phase, the agent is trained on a large-scale dataset $\mathcal{D}_{\text{pretrain}}$ consisting of diverse indoor scenes.
Each training sample is a tuple $(s_0, g_{\text{img}}, \tau)$, where $s_0$ denotes the initial state, $g_{\text{img}} \in \mathbb{R}^{H \times W \times 3}$ is a goal image specifying the target location, and $\tau = \{(s_t, a_t)\}_{t=0}^{T}$ is a trajectory of state-action pairs leading to the goal.

The agent learns a parameterized policy $\pi_{\text{base}}(a_t \mid s_t)$. Here, the observation $s_t = \{o_t, g_{\text{img}}, \mathcal{O}_{<t}\}$ at time $t$ consists of the current RGB frame $o_t \in \mathbb{R}^{H \times W \times 3}$, the goal image $g_{\text{img}}$, and the stack of all historical RGB frames $\mathcal{O}_{<t} = \{o_0, \ldots, o_{t-1}\}$, without including proprioceptive or low-level odometry states.
The action $a_t \in \mathbb{R}^{K \times 3}$ is defined as a continuous waypoint sequence that specifies $K$ target positions in the robot's egocentric coordinate frame.

The pretraining objective is to minimize the expected imitation loss over the dataset:
\begin{equation}
\mathcal{L}_{\text{pretrain}} = \mathbb{E}_{(s_0, g_{\text{img}}, \tau) \sim \mathcal{D}_{\text{pretrain}}} \left[ \sum_{t=0}^{T} \ell\big(\pi_{\text{base}}(s_t), a_t^*\big) \right],
\end{equation}
where $\ell(\cdot, \cdot)$ is the mean squared error between the predicted and ground-truth waypoint sequences, and $a_t^*$ denotes the continuous expert action at time $t$.

\subsubsection{Online Self-Learning Phase}

After pretraining, the agent is deployed in a new environment $E$ and enters the online self-learning phase.
The agent receives no further human supervision and must autonomously improve its navigation capabilities through interaction with $E$.

\paragraph{Input and Setup}
The process begins with the pretrained policy $\pi_{\text{base}}$ parameterized by $\theta_0$ and an external internet image collection $\mathcal{G}_{\text{internet}} = \{g_i^{\text{net}}\}_{i=1}^{N_{\text{net}}}$, where each $g_i^{\text{net}}$ is an RGB image of common indoor objects or scenes retrieved from large-scale web sources.
At deployment time, the agent receives arbitrary open-vocabulary natural language instructions $I$ from users.

\paragraph{Process and Output}
The agent interacts with environment $E$ over a series of episodes $k = 1, 2, \ldots$.
During each episode, the agent first explores the environment by executing actions sampled from the current policy $\pi_{\theta_k}$.
Simultaneously, it captures observations $\{o_t^{(k)}\}_{t=1}^{T_k}$ along its trajectory and stores them in a persistent multimodal memory database $\mathcal{M}_k$.
The agent then updates the memory by adding new observations, generating textual descriptions, and removing redundant entries to maintain diversity.
To drive learning, the agent generates self-supervised training tasks by sampling instructions $I_j$ from the memory content and retrieving corresponding goal images $g_j$ either from $\mathcal{M}_k$ or from $\mathcal{G}_{\text{internet}}$.
Finally, the policy $\pi_{\theta_k}$ is optimized using reinforcement learning based on these self-generated tasks and a fixed similarity-based reward function that measures visual alignment to the goal.

Formally, at episode $k$, the memory database is represented as:
\begin{equation}
\mathcal{M}_k = \left\{ (o_i, d_i, \tau_i) \mid i = 1, \ldots, N_k \right\},
\end{equation}
where each entry consists of an RGB keyframe image $o_i \in \mathbb{R}^{H \times W \times 3}$, a textual description $d_i \in \mathcal{V}^*$ generated by a vision-language model (with $\mathcal{V}$ being the vocabulary), and a timestamp $\tau_i \in \mathbb{R}_+$ indicating when the observation was captured.

The policy parameters $\theta_k$ evolve over episodes according to a reinforcement learning update:
\begin{equation}
\theta_{k+1} = \theta_k + \alpha \nabla_\theta \mathbb{E}_{\pi_{\theta_k}} \left[ \sum_{t=0}^{T} \gamma^t r_t \right],
\end{equation}
where $\alpha$ is the learning rate, $\gamma \in (0,1)$ is the discount factor, and $r_t$ is the reward at time $t$, computed based on the visual similarity between the current observation $o_t$ and the goal image $g$.

\subsection{Core Constraints and Challenges}

The lifelong self-learning embodied navigation task is defined by three strict constraints that introduce significant technical challenges.

First, the process is \textbf{strictly unsupervised}: the agent receives no human-provided labels, demonstrations, or reward annotations during the online phase.
This imposes the challenge of \emph{self-driven curriculum learning}, where the agent must autonomously generate diverse and informative training tasks (instruction-goal pairs) to drive policy optimization without external guidance.

Second, we enforce a \textbf{zero instruction prior} setting, where the test-time instructions are disjoint from any seen during pretraining.
To support continuous improvement, the system must address the \emph{exploration challenge}, effectively aligning static internet priors with dynamic, fragmented embodied observations to motivate the agent to discover and map new areas beyond its current experience.

Third, the learning is \textbf{online and incremental}.
The agent operates on a continuous data stream, requiring a \emph{persistent memory mechanism} that consolidates fragmented views into a coherent global understanding.
This necessitates solving the stability-plasticity dilemma: adapting to environmental changes and new concepts without catastrophically forgetting previously acquired skills.

%% file: sections/4_method.tex
\section{The \ours Framework}

This section presents \ours, a lifelong self-learning embodied navigation framework that autonomously acquires open-vocabulary navigation capabilities through continuous interaction with dynamic environments.
We begin with a high-level overview of the system architecture, then detail each core component: the self-evolving multimodal memory database, the self-instruction generation and hybrid goal retrieval mechanism, the pretrained VLA navigation backbone, and the conservative Q-learning-based policy optimization module.
We conclude with implementation details that enable robust sim-to-real transfer and efficient online learning.

\subsection{Framework Overview}

\begin{algorithm}[!t]
\caption{Lifelong Self-Learning Embodied Navigation}
\label{alg:method}
\begin{algorithmic}[1]
\REQUIRE Pretrained VLA Policy $\pi_\theta$, VLM, Internet Goal Database $\mathcal{G}_{\text{internet}}$
\STATE \textbf{Initialize:} Memory Database $\mathcal{M} \leftarrow \emptyset$, Replay Buffer $\mathcal{B} \leftarrow \emptyset$, Critic $Q_\phi$
\STATE \textbf{Start Parallel Processes:}

\STATE \textbf{Process 1: Active Exploration Loop}
\LOOP
    \STATE \textbf{// Self-Instruction Generation}
    \IF{update condition met (period or $|\mathcal{M}|$ growth)}
        \STATE Construct context $\mathcal{C} \leftarrow \{d_i \mid (o_i, d_i) \in \mathcal{M}\}$
        \STATE Generate task set $\{I_j\} \leftarrow \text{VLM}_{\text{generate}}(\mathcal{C})$
    \ENDIF
    
    \STATE \textbf{// Hybrid Goal Retrieval}
    \STATE Sample instruction $I \sim \{I_j\}$
    \STATE Retrieve from memory: $g_{\text{mem}}, c \leftarrow \text{Retrieve}(I, \mathcal{M})$
    \IF{$c < \tau_{\text{conf}}$ \OR random $< p_{\text{fallback}}$}
        \STATE Fallback: $g \leftarrow \text{Retrieve}(I, \mathcal{G}_{\text{internet}})$
    \ELSE
        \STATE Set goal: $g \leftarrow g_{\text{mem}}$
    \ENDIF
    
    \STATE \textbf{// Online Exploration \& Memory Update}
    \STATE Initialize episode $t \leftarrow 0$
    \WHILE{not done}
        \STATE Observe $o_t$, construct state $s_t = (o_t, g, \mathcal{H}_t)$
        \STATE Select action $a_t \sim \pi_\theta(\cdot|s_t)$
        \STATE Execute $a_t$, observe reward $r_t$ and next state $s_{t+1}$
        \STATE Store transition $(s_t, a_t, r_t, s_{t+1})$ in $\mathcal{B}$
        \STATE \textbf{// Memory Update}
        \IF{$o_t$ is novel or offers better alignment}
            \STATE Update $\mathcal{M}$ with $(o_t, \text{VLM}(o_t))$
        \ENDIF
        \STATE $t \leftarrow t + 1$
    \ENDWHILE
\ENDLOOP

\STATE \textbf{Process 2: Policy Optimization Loop}
\LOOP
    \STATE \textbf{// Continuous Policy Optimization}
    \IF{$|\mathcal{B}| > \text{batch size}$}
        \STATE Sample batch from $\mathcal{B}$
        \STATE Update Critic $Q_\phi$ by minimizing $\mathcal{L}_{\text{CQL}}$
        \STATE Update Policy $\pi_\theta$ by minimizing $\mathcal{L}_{\text{policy}}$
    \ENDIF
\ENDLOOP
\end{algorithmic}
\end{algorithm}

\begin{figure*}[!t]
\centering
\includegraphics[width=0.95\textwidth]{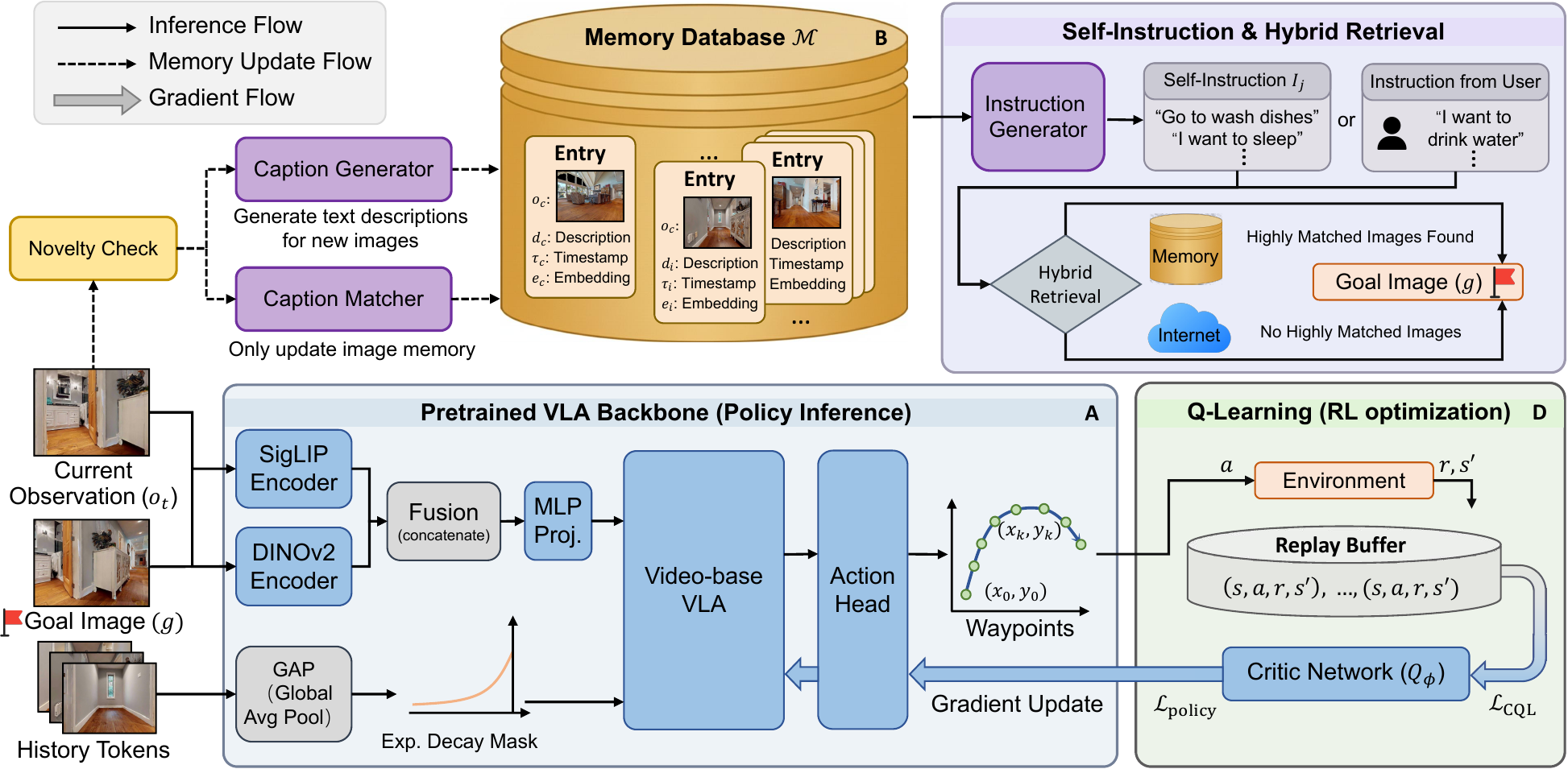}
\caption{\ours system architecture. \textbf{(A) The VLA Navigation Backbone} processes dual-encoded observations (SigLIP + DINOv2) and compressed history to predict waypoints. \textbf{(B) The Self-Evolving Memory Database} continuously accumulates keyframes with semantic descriptions generated by VLM. \textbf{(C) The Self-Instruction \& Retrieval Module} generates diverse tasks from memory and retrieves visual goals (with Internet fallback). \textbf{(D) The CQL-based RL Module} refines the policy using a conservative objective to ensure stability.}
\label{fig:pipeline}
\end{figure*}

\ours implements a closed-loop learning system driven by a continuously evolving multimodal memory database.
The system architecture, illustrated in Figure~\ref{fig:pipeline}, comprises four integrated modules that collectively realize the memory--policy co-evolution cycle.

At the heart of \ours is a vision-language-action (VLA) model $\pi_\theta$ that maps observations and goal images to navigation actions.
The model is initialized through pretraining on large-scale image-goal navigation datasets, acquiring foundational visuomotor skills and spatial reasoning capabilities.
Upon deployment, the agent autonomously explores the environment while constructing a persistent memory database $\mathcal{M}$ that encodes visual appearances, semantic descriptions, and temporal context.

The memory facilitates two critical functions.
First, it supports \textbf{self-supervised task generation}, where the system periodically samples textual descriptions from $\mathcal{M}$ and feeds them to a language model to generate diverse, contextually relevant navigation instructions $\{I_j\}$.
For each instruction, a hybrid retrieval mechanism first searches the memory database for matching goal images; if no suitable match exists, it falls back to an external internet image collection $\mathcal{G}_{\text{internet}}$, ensuring continuous availability of training signals even during early exploration.
Second, it enables \textbf{instruction-conditioned navigation at test time}.
When a user issues an open-vocabulary instruction $I$, the system retrieves the most semantically relevant image $g$ from $\mathcal{M}$ or $\mathcal{G}_{\text{internet}}$ and uses it as the goal for the VLA policy, enabling zero-shot generalization to novel instructions.

Policy improvement is driven by a conservative Q-learning (CQL) objective that mitigates Q-value overestimation in offline settings while maintaining stability during online updates.
The reward signal is derived from a learned visual similarity function that measures alignment between the agent's current observation and the goal image, providing both terminal success rewards and intermediate progress signals to guide exploration.

As the agent accumulates experience, the memory grows and diversifies, enabling progressively more challenging self-generated tasks.
This virtuous cycle—exploration enriches memory, memory enables better task generation, tasks drive policy improvement, and improved policy enables more efficient exploration—forms the foundation of \ours's lifelong learning capability. The complete training procedure is summarized in Algorithm~\ref{alg:method}.

\subsection{Pretrained VLA Navigation Backbone}

The navigation policy is built upon a transformer-based vision-language-action model that jointly processes egocentric visual observations and goal images to predict navigation actions.
We describe the visual encoding and observation sequence construction below, and detail the pretraining strategy in a separate subsection.

\subsubsection{Multimodal Visual Encoding}

To capture both semantic and structural information from RGB observations, we employ a dual-encoder architecture that combines complementary visual representations.
Concretely, for each RGB frame $o_t \in \mathbb{R}^{H \times W \times 3}$, we extract features using two pretrained vision transformers:
\begin{equation}
\begin{aligned}
\mathbf{f}_t^{\text{sem}} &= \text{SigLIP}(o_t) \in \mathbb{R}^{N_{\text{patch}} \times d_{\text{sem}}}, \\
\mathbf{f}_t^{\text{struct}} &= \text{DINOv2}(o_t) \in \mathbb{R}^{N_{\text{patch}} \times d_{\text{struct}}},
\end{aligned}
\end{equation}
where SigLIP~\cite{zhai2023sigmoid} provides language-aligned semantic representations trained on large-scale image-text pairs, and DINOv2~\cite{oquab2023dinov2} captures fine-grained structural and spatial details through self-supervised learning.
We use SigLIP and DINOv2, which operate on different input resolutions to extract complementary visual features.

We then perform feature fusion and projection, where the complementary features are concatenated along the channel dimension and projected into the VLA model's embedding space via a learned multimodal projector:
\begin{equation}
\mathbf{z}_t = \text{MLP}_{\text{proj}}\left( \left[ \mathbf{f}_t^{\text{sem}} \, ; \, \mathbf{f}_t^{\text{struct}} \right] \right) \in \mathbb{R}^{N_{\text{patch}} \times d_{\text{model}}},
\end{equation}
where $[\cdot ; \cdot]$ denotes concatenation, and $\text{MLP}_{\text{proj}}$ is a 2-layer MLP with GELU activation and layer normalization.
This design enables the model to leverage both high-level semantic understanding for goal recognition and low-level geometric cues for obstacle avoidance and spatial reasoning.

\subsubsection{Observation Sequence Construction with History Compression}

To provide the VLA model with temporal context while maintaining computational efficiency, we construct a token sequence that balances detailed representation of critical frames with compact encoding of historical observations.

We encode the current observation $o_t$ and the goal image $g$ with full spatial resolution, preserving detailed visual information:
\begin{equation}
\mathbf{Z}_{\text{curr}} = \mathbf{z}_t \in \mathbb{R}^{64 \times d_{\text{model}}}, \quad \mathbf{Z}_{\text{goal}} = \mathbf{z}_g \in \mathbb{R}^{64 \times d_{\text{model}}}.
\end{equation}

For computational efficiency and lifelong context, we compress historical observations using Global Average Pooling (GAP).
Each historical frame $o_{t'}$ for $t' < t$ is spatially pooled:
\begin{equation}
\mathbf{z}_{t'}^{\text{pool}} = \text{GAP}(\mathbf{z}_{t'}),
\end{equation}
where GAP applies simple average pooling to aggregate spatial features.
This reduces the token count per historical frame, enabling the model to attend to a longer temporal window within a fixed computational budget.

To prioritize recent observations while maintaining awareness of distant history, we apply a temporal attention bias:
\begin{equation}
\alpha_{t, t'} = \exp\left( -\lambda (t - t') \right) \quad \text{for } t' < t,
\end{equation}
where $\lambda > 0$ controls the decay rate.
This bias is added to the attention logits before softmax, ensuring that older frames contribute less to the model's decision-making without being entirely discarded.

The complete input to the VLA transformer at time $t$ is:
\begin{equation}
\mathbf{X}_t = \left[ \mathbf{Z}_{\text{goal}} \, ; \, \mathbf{Z}_{\text{curr}} \, ; \, \{\mathbf{z}_{t'}^{\text{pool}}\}_{t' \in \mathcal{H}_t} \right],
\end{equation}
where $\mathcal{H}_t$ denotes the set of historical frames selected based on the exponential decay attention mask, prioritizing recent observations while maintaining awareness of distant history.
The total token count is $64 + 64 + N_{\text{hist}} \times 4$ tokens, where $N_{\text{hist}} = |\mathcal{H}_t|$ is the number of historical frames actually included in the input sequence, determined by the attention mask filtering mechanism.
We set the history window size to 64 frames.

\subsubsection{Action Representation and Prediction}

Rather than predicting instantaneous velocities, we adopt a waypoint-based action representation that facilitates lifelong planning and smooth trajectory execution.
At each timestep, the model predicts a sequence of $K=8$ future waypoints in the robot's local coordinate frame:
\begin{equation}
\mathbf{a}_t = \left\{ (\Delta x_k, \Delta y_k, \Delta \theta_k) \right\}_{k=1}^{K} \in \mathbb{R}^{K \times 3},
\end{equation}
where $(\Delta x_k, \Delta y_k)$ specifies the relative position of the $k$-th waypoint and $\Delta \theta_k$ is the relative heading.
This representation provides a smooth motion plan that the low-level controller can execute robustly.

The VLA transformer output is passed through a 4-layer MLP action head with residual connections:
\begin{equation}
\mathbf{a}_t = \text{MLP}_{\text{action}}\left( \text{Transformer}_\theta(\mathbf{X}_t) \right),
\end{equation}
where $\text{Transformer}_\theta$ denotes the pretrained Qwen2-7B backbone with learned visual adapters.

\subsection{Self-Evolving Multimodal Memory Database}

The memory database $\mathcal{M}$ serves as the agent's persistent internal representation of the environment, supporting both semantic retrieval for instruction-conditioned navigation and self-supervised task generation for policy learning.
Unlike static scene representations, $\mathcal{M}$ evolves continuously as the agent explores, incorporating new observations while maintaining diversity and preventing redundancy.

\subsubsection{Memory Entry Structure}

Each entry in $\mathcal{M}$ is a quadruple $(o_i, d_i, \tau_i, \mathbf{e}_i)$.
The first component, $o_i \in \mathbb{R}^{H \times W \times 3}$, is an RGB visual keyframe captured during exploration.
The second component, $d_i \in \mathcal{V}^*$, is a natural language semantic description of the scene content.
The third component, $\tau_i \in \mathbb{R}_+$, represents the timestamp for temporal reasoning and adaptation to dynamic environments.
Finally, $\mathbf{e}_i \in \mathbb{R}^{d_{\text{embed}}}$ is the precomputed visual embedding extracted using LLaVE~\cite{lan2025llave}, which facilitates efficient similarity search and retrieval without repeated inference.

\subsubsection{Autonomous Memory Construction}

During exploration, the agent captures observations at regular intervals (every 5 steps).
A frame is identified as a candidate keyframe if its visual embedding deviates significantly from the previously selected keyframe:
\begin{equation}
\Delta_t = 1 - \frac{\mathbf{e}_t \cdot \mathbf{e}_{\text{last}}}{\|\mathbf{e}_t\| \|\mathbf{e}_{\text{last}}\|} > \delta_{\text{change}},
\end{equation}
where $\delta_{\text{change}} = 0.15$ and $\mathbf{e}_{\text{last}}$ denotes the embedding of the most recently added keyframe.
This filtering step efficiently identifies visually distinct moments in the trajectory for potential inclusion in the memory, independently of the global memory content.

\subsubsection{Memory Update and Deduplication}

To maintain a compact and high-quality memory database, we evaluate each candidate keyframe $o_{\text{new}}$ against the existing entries in $\mathcal{M}$.
We first identify the most visually similar existing entry:
\begin{equation}
j = \arg\max_i \left( s_i = \frac{\mathbf{e}_{\text{new}} \cdot \mathbf{e}_i}{\|\mathbf{e}_{\text{new}}\| \|\mathbf{e}_i\|} \right).
\end{equation}

If the maximum similarity $s_j \le \tau_{\text{dup}}$ (where $\tau_{\text{dup}} = 0.7$), the scene is considered novel.
We invoke Qwen3-VL-8B-Instruct to generate a semantic description $d_{\text{new}} = \text{VLM}_{\text{caption}}(o_{\text{new}})$ and insert the new entry:
\begin{equation}
\mathcal{M} \leftarrow \mathcal{M} \cup \{(o_{\text{new}}, d_{\text{new}}, \tau_{\text{new}}, \mathbf{e}_{\text{new}})\}.
\end{equation}

If $s_j > \tau_{\text{dup}}$, the candidate overlaps with an existing entry.
We update the entry only if the new image provides a better visual grounding for the stored semantic concept.
Specifically, we compare the alignment between the image embeddings and the text embedding of the existing description $d_j$, computed using the LLaVE text encoder $\mathcal{E}_{\text{txt}}$:
\begin{equation}
\text{align}_{\text{new}} = \mathbf{e}_{\text{new}} \cdot \mathcal{E}_{\text{txt}}(d_j), \quad \text{align}_{\text{old}} = \mathbf{e}_j \cdot \mathcal{E}_{\text{txt}}(d_j).
\end{equation}
If $\text{align}_{\text{new}} > \text{align}_{\text{old}}$, we replace the visual component of the entry:
\begin{equation}
\mathcal{M} \leftarrow (\mathcal{M} \setminus \{(o_j, d_j, \tau_j, \mathbf{e}_j)\}) \cup \{(o_{\text{new}}, d_j, \tau_{\text{new}}, \mathbf{e}_{\text{new}})\}.
\end{equation}
This ensures that the memory maintains the most representative visual instance for each scene description.

\begin{figure*}[!t]
\centering
\includegraphics[width=0.95\textwidth]{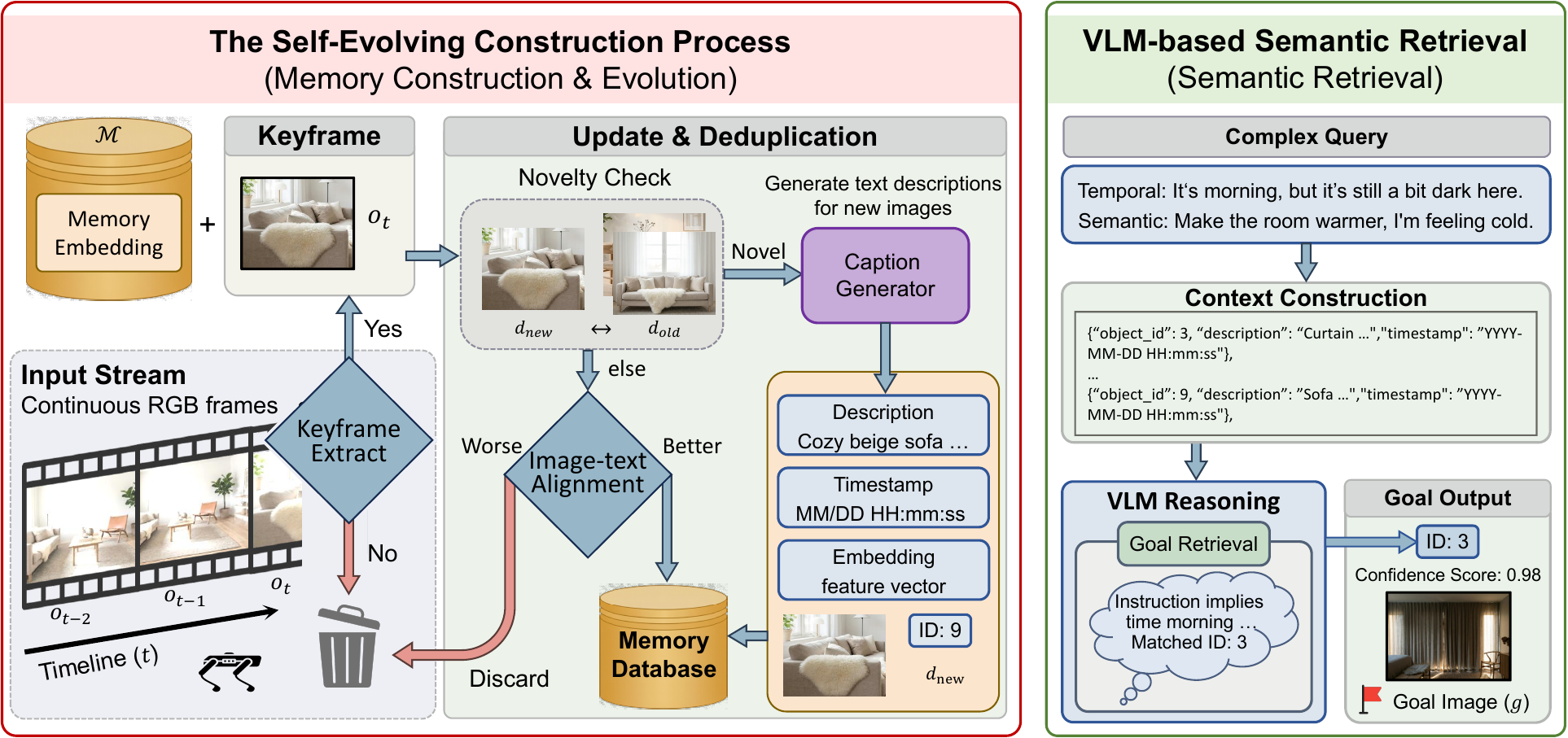}
\caption{Memory update and retrieval mechanism. \textbf{(Left) The Self-Evolving Construction Process:} Novel frames are captioned and inserted; redundant entries trigger an alignment-based update check. \textbf{(Right) VLM-Based Semantic Retrieval:} Given a complex instruction (possibly with temporal constraints), the VLM selects the most relevant goal from a context of candidate descriptions and timestamps.}
\label{fig:memory}
\vspace{-0.3cm}
\end{figure*}

Figure~\ref{fig:memory} details the construction and retrieval processes of the memory database.

\subsubsection{VLM-Based Semantic Retrieval}

Given a natural language instruction $I$, we retrieve the most relevant visual goal from $\mathcal{M}$ by directly leveraging the semantic reasoning capabilities of the Vision-Language Model.
Instead of relying on embedding-based similarity search, we formulate retrieval as a selection task for Qwen3-VL-8B-Instruct.
We construct a context containing the current time $t_{\text{curr}}$, the instruction, and the set of all available scene descriptions with their timestamps:
\begin{equation}
\begin{aligned}
P_{\text{ret}} = &\text{``Current Time: } t_{\text{curr}}. \text{ Given the instruction: } I, \\
&\text{select the best matching scene from: } \{(d_i, \tau_i)\}_{i=1}^{|\mathcal{M}|}\text{''}.
\end{aligned}
\end{equation}
The inclusion of timestamps $\tau_i$ enables the VLM to handle time-sensitive instructions (e.g., ``Go to the place visited 5 minutes ago'' or ``Find the office seen this morning''), crucial for dynamic environment adaptation.
The model returns the index $i^*$ of the best matching entry along with a confidence score $c^* \in [0, 1]$:
\begin{equation}
(i^*, c^*) = \text{VLM}_{\text{select}}(P_{\text{ret}}), \quad g = o_{i^*}.
\end{equation}
where $c^*$ represents the model's confidence in the retrieval result, derived from the probability of the generated token.
This approach allows the system to handle complex, compositional, and functional instructions that simple embedding dot-products often fail to capture.

\subsection{Self-Instruction Generation and Hybrid Goal Retrieval}

To enable fully autonomous learning without human supervision, \ours must generate its own training tasks in the form of instruction-goal pairs.
We develop a self-instruction mechanism that produces diverse, contextually relevant navigation instructions and a hybrid retrieval strategy that ensures continuous availability of training signals.

\subsubsection{Contextual Self-Instruction Generation}

Periodically (every 10 episodes or when $|\mathcal{M}|$ increases by 20\%), we invoke Qwen3-VL-8B-Instruct to generate navigation instructions based on the accumulated memory content.

We aggregate all semantic descriptions in the memory database:
\begin{equation}
\mathcal{C}_{\text{memory}} = \left\{ d_i \mid (o_i, d_i, \tau_i, \mathbf{e}_i) \in \mathcal{M} \right\},
\end{equation}
and construct a prompt:
\begin{equation}
\begin{aligned}
P_{\text{gen}} = &\text{``Based on the following scene descriptions: } \\
&[\mathcal{C}_{\text{memory}}], \\
&\text{generate 50 diverse navigation instructions''}.
\end{aligned}
\end{equation}

The VLM then generates a list of open-vocabulary instructions:
\begin{equation}
\{I_j\}_{j=1}^{M} = \text{VLM}_{\text{generate}}(P_{\text{gen}}),
\end{equation}
where $M = 50$ instructions are produced per generation round.
The instructions span multiple difficulty levels and semantic categories to ensure robust training.
These include \emph{object-centric} tasks (e.g., ``Go to the red sofa'', ``Find the coffee machine''), \emph{functional} queries (e.g., ``I need to wash my hands'', ``Where can I charge my phone?''), \emph{spatial} directives (e.g., ``Navigate to the room with the large window'', ``Go to the corner with plants''), \emph{compositional} instructions (e.g., ``Find a quiet place to read'', ``Locate the meeting area''), and \emph{temporal} instructions (e.g., ``Go to the room visited 10 minutes ago'', ``Find the location seen at the start of exploration'').
At the beginning of each new episode, one instruction is randomly sampled from the current set $\{I_j\}$ to guide the agent's exploration and learning.

This diversity ensures that the policy is trained on a broad spectrum of tasks, promoting generalization.

\subsubsection{Hybrid Goal Retrieval Mechanism}

For each generated instruction $I_j$, we must identify a corresponding goal image $g_j$ that the agent can navigate to.
We employ a hybrid retrieval strategy that prioritizes memory-based goals while falling back to internet images when necessary.

We first attempt to retrieve a goal from the memory database $\mathcal{M}$ using the VLM-based retrieval procedure described in Section 4.3.4:
\begin{equation}
g_j^{\text{memory}}, c_j^{\text{memory}} = \text{Retrieve}(I_j, \mathcal{M}).
\end{equation}

To ensure robust training signals, we employ a probabilistic fallback mechanism.
If the memory retrieval confidence $c_j^{\text{memory}}$ is below a threshold $\tau_{\text{conf}} = 0.6$, we trigger a fallback to the external internet collection $\mathcal{G}_{\text{internet}}$.
Additionally, during early exploration when $\mathcal{M}$ is sparse, we fallback with a probability that decays as memory grows:
\begin{equation}
p_{\text{fallback}} = \max\left(0, 1 - \frac{|\mathcal{M}|}{20}\right).
\end{equation}
If fallback is triggered (either due to low confidence or the probabilistic sampling), we retrieve a goal image from the external internet collection:
\begin{equation}
g_j^{\text{internet}} = \text{Retrieve}(I_j, \mathcal{G}_{\text{internet}}),
\end{equation}
where $\mathcal{G}_{\text{internet}}$ is a diverse collection of indoor scene images crawled from the internet.

The final goal for instruction $I_j$ is determined by the outcome of the fallback logic:
\begin{equation}
g_j = 
\begin{cases}
g_j^{\text{internet}} & \text{if fallback triggered}, \\
g_j^{\text{memory}} & \text{otherwise}.
\end{cases}
\end{equation}

This hybrid strategy ensures that the agent trains primarily on goals that exist in the current environment (via $\mathcal{M}$), maximizing training efficiency.
Simultaneously, it prevents training from stalling during early exploration when the memory is sparse by leveraging $\mathcal{G}_{\text{internet}}$.
Furthermore, the inclusion of internet images motivates the agent to explore unseen regions and novel concepts that have not yet been captured in the episodic memory, thereby driving continuous discovery.

\subsection{Pretraining Strategy and Data Augmentation}

We pretrain the VLA model on a large-scale dataset $\mathcal{D}_{\text{pretrain}}$ collected in the Habitat simulator using scenes from HM3D~\cite{ramakrishnan2021habitat} and MP3D~\cite{chang2017matterport3d}.
The dataset comprises 50,000 trajectories totaling 5 million steps, spanning diverse indoor environments, with a carefully balanced mix.
Specifically, 80\% of the data consists of expert trajectories representing shortest-path navigation from random start positions to sampled goal images, providing efficient navigation demonstrations.
The remaining 20\% comprises exploratory trajectories formed by random walks with periodic goal-switching, which encourages exploration and enhances robustness to suboptimal paths.

To bridge the sim-to-real gap, we apply aggressive image augmentations independently to the current frame $o_t$ and the goal frame $g$, artificially increasing appearance variation:
\begin{equation}
\begin{aligned}
\tilde{o}_t &= \mathcal{A}_{\text{aug}}(o_t; \xi_t), \\
\tilde{g} &= \mathcal{A}_{\text{aug}}(g; \xi_g),
\end{aligned}
\end{equation}
where $\mathcal{A}_{\text{aug}}$ includes motion blur (kernel size 3--7), brightness/contrast jitter ($\pm 30\%$), color jitter (saturation $\pm 20\%$), and Gaussian noise ($\sigma \in [0, 0.1]$).
To account for quadruped locomotion and sensor artifacts, we further introduce: (1) random camera shake to simulate head motion; (2) random lighting variation; (3) random color block insertion; and (4) random region removal, where the removed area is smoothly filled inwards using boundary pixel colors.
The independent sampling $\xi_t \neq \xi_g$ ensures that the model learns to match goals under varying visual conditions, improving robustness to real-world appearance shifts.

We train the model via supervised learning with mean squared error (MSE) loss on waypoint predictions:
\begin{equation}
\mathcal{L}_{\text{pretrain}} = \mathbb{E}_{(o_t, g, \mathbf{a}_t^*) \sim \mathcal{D}_{\text{pretrain}}} \left[ \| \pi_\theta(o_t, g, \mathcal{O}_{<t}) - \mathbf{a}_t^* \|_2^2 \right],
\end{equation}
where $\mathbf{a}_t^*$ is the ground-truth waypoint sequence computed from the expert trajectory.
Training is performed for 14K steps with batch size 320, using AdamW optimizer with learning rate $3.3 \times 10^{-5}$ and cosine annealing schedule.

\begin{figure*}[!t]
\centering
\vspace{0.2cm}
\includegraphics[width=0.95\textwidth]{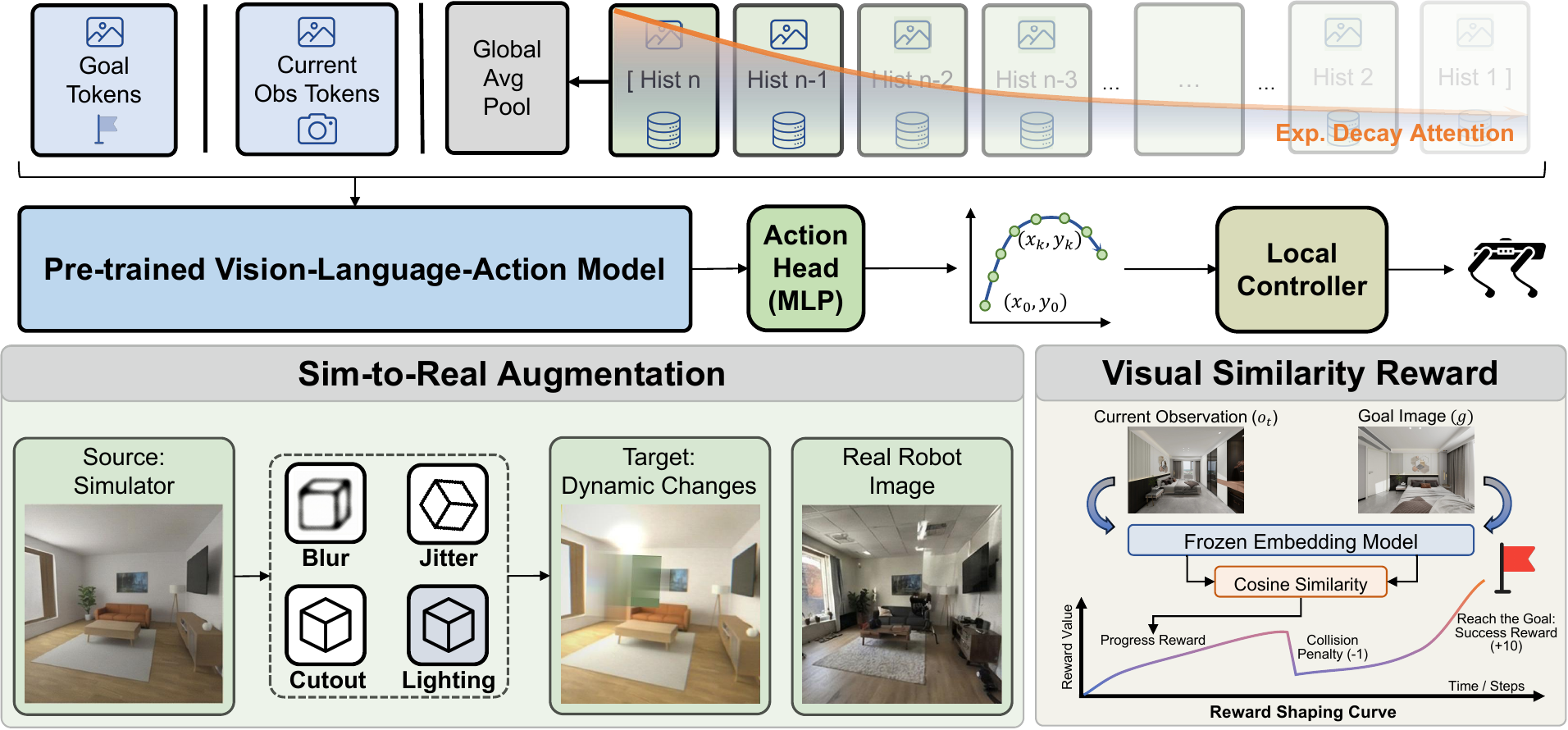}
\caption{VLA backbone architecture and Sim-to-Real strategy. \textbf{(Top)} The architecture processes goal tokens, current observation tokens, and compressed history tokens (weighted by an exponential decay mask) through a Qwen2-7B backbone. \textbf{(Bottom)} We apply aggressive visual augmentations (blur, noise, cutout) during pretraining to bridge the domain gap. The model outputs 8 local waypoints as a robust action representation.}
\label{fig:vla}
\end{figure*}

Figure~\ref{fig:vla} illustrates the complete VLA architecture, showing the flow from dual visual encoders through the transformer backbone to waypoint prediction.

\subsection{Conservative Q-Learning for Lifelong Policy Optimization}

To refine the pretrained navigation policy through online interaction, we employ reinforcement learning with conservative Q-learning (CQL).
CQL provides a principled framework for learning from mixed-quality data (both online rollouts and offline pretraining trajectories) while mitigating the Q-value overestimation problem that plagues standard off-policy RL algorithms.

\subsubsection{Markov Decision Process Formulation}

We model the navigation task as an infinite-horizon MDP $(\mathcal{S}, \mathcal{A}, \mathcal{T}, r, \gamma)$.
Here, $\mathcal{S}$ denotes the state space, with $s_t = \{o_t, g, \mathcal{O}_{<t}\}$ encoding the current RGB observation, goal image, and historical RGB frames.
The continuous action space $\mathcal{A} = \mathbb{R}^{K \times 3}$ consists of waypoint sequences.
The environment transition dynamics are given by $\mathcal{T}(s_{t+1} \mid s_t, a_t)$, while $r: \mathcal{S} \times \mathcal{A} \to \mathbb{R}$ represents the reward function (detailed below), and $\gamma = 0.99$ serves as the discount factor.

The goal is to learn a policy $\pi_\theta: \mathcal{S} \to \mathcal{A}$ that maximizes expected cumulative return:
\begin{equation}
J(\theta) = \mathbb{E}_{\pi_\theta} \left[ \sum_{t=0}^{\infty} \gamma^t r(s_t, a_t) \right].
\end{equation}

\subsubsection{Reward Function Design}

The reward function combines terminal success rewards, intermediate progress signals, and regularization terms to guide efficient and collision-free navigation.

We leverage the LLaVE embeddings to compute the semantic similarity between the current observation $o_t$ and the goal image $g$:
\begin{equation}
\text{sim}_t = \frac{\mathbf{e}_t \cdot \mathbf{e}_g}{\|\mathbf{e}_t\| \|\mathbf{e}_g\|}.
\end{equation}
This metric provides a robust measure of semantic alignment for reward calculation.
To ensure that reward computation does not throttle the control loop, LLaVE inference is executed asynchronously on a parallel stream, decoupling the reward signal generation from the real-time VLA policy execution.

The total reward at time $t$ is defined as the sum of four components:
\begin{equation}
r_t = r_t^{\text{term}} + r_t^{\text{prog}} + r_t^{\text{time}} + r_t^{\text{coll}}.
\end{equation}

The \emph{terminal reward} $r_t^{\text{term}}$ incentivizes successful completion:
\begin{equation}
r_t^{\text{term}} = 
\begin{cases}
+10.0 & \text{if agent stops and } \text{sim}_t \geq 0.75, \\
0 & \text{otherwise}.
\end{cases}
\end{equation}
Empirically, a similarity score above 0.75 indicates a high probability that the agent has reached the target location. We verify this in our experiments by comparing against human judgment of goal arrival.

To mitigate the issue of sparse rewards, we include a \emph{progress reward} $r_t^{\text{prog}}$.
Since visual similarity scores below 0.5 typically indicate negligible semantic overlap between the current view and the goal, fluctuations in this range do not trigger rewards.
However, once the similarity exceeds 0.5, it signifies that the agent has entered the vicinity of the target.
To encourage continued progress, we assign a bonus of $+2.0$ whenever the similarity score $\text{sim}_t$ surpasses the thresholds $\{0.5, 0.6, 0.7\}$ for the first time in an episode.

Finally, we apply a \emph{time penalty} $r_t^{\text{time}} = -0.01$ at each step to encourage efficiency, and a \emph{collision penalty} $r_t^{\text{coll}} = -0.1$ whenever a collision is detected to ensure safety.

\subsubsection{Conservative Q-Learning Objective}

Standard actor-critic RL methods suffer from Q-value overestimation, particularly when learning from offline data or mixed online-offline experiences.
CQL addresses this by penalizing Q-values for out-of-distribution actions, encouraging the critic to remain conservative.

We train a separate MLP-based critic network $Q_\phi(s, a)$ with parameters $\phi$, consisting of 4 hidden layers with ReLU activations:
\begin{equation}
Q_\phi(s_t, a_t) = \text{MLP}_\phi\left( \left[ s_t \, ; \, a_t \right] \right),
\end{equation}
where $[\cdot ; \cdot]$ denotes concatenation of the state and the action.

The CQL objective augments the standard Bellman error with a conservative regularization term:
\begin{equation}
\begin{aligned}
\mathcal{L}_{\text{CQL}}(\phi) = &\mathbb{E}_{(s, a, r, s') \sim \mathcal{B}} \Big[ \Big( Q_\phi(s, a) \\
&\quad - \left( r + \gamma \max_{a'} Q_{\phi'}(s', a') \right) \Big)^2 \Big] \\
&+ \alpha \mathbb{E}_{s \sim \mathcal{B}} \left[ \log \sum_{a \sim \mu(a \mid s)} \exp(Q_\phi(s, a)) \right] \\
&- \alpha \mathbb{E}_{s \sim \mathcal{B}, a \sim \pi_\theta(a \mid s)} [Q_\phi(s, a)],
\end{aligned}
\end{equation}
In this objective, $\mathcal{B}$ represents the replay buffer storing transitions from online exploration, while $\phi'$ denotes the parameters of a target Q-network updated via exponential moving average.
The second term serves as a conservative regularizer, penalizing high Q-values for out-of-distribution actions sampled from a uniform random policy $\mu(a \mid s)$.
By minimizing the difference between the log-sum-exp of Q-values under $\mu$ and the expected Q-value under the current policy $\pi_\theta$, this term effectively lower-bounds the Q-function and prevents overestimation, with $\alpha = 5.0$ controlling the regularization strength.

The policy optimization objective combines Q-value maximization with a behavior cloning (BC) regularization term to ensure stability:
\begin{equation}
\begin{aligned}
\mathcal{L}_{\text{policy}}(\theta) = &-\mathbb{E}_{s \sim \mathcal{B}, a \sim \pi_\theta(\cdot \mid s)} \left[ Q_\phi(s, a) \right] \\
&+ \lambda_{\text{bc}} \mathbb{E}_{(s, a^*) \sim \mathcal{B}_{\text{succ}}} \left[ \| \pi_\theta(s) - a^* \|_2^2 \right],
\end{aligned}
\end{equation}
where $\mathcal{B}_{\text{succ}} \subset \mathcal{B}$ denotes the subset of transitions from successful trajectories, and $\lambda_{\text{bc}} = 0.1$ weighs the contribution of the imitation loss.
The BC term acts as a stabilizer, encouraging the policy to stay close to known successful actions while the Q-term drives improvement.

We alternate between critic updates (minimizing $\mathcal{L}_{\text{CQL}}$) and policy updates (minimizing $\mathcal{L}_{\text{policy}}$) using gradient descent, with updates performed every 100 steps.

\subsubsection{Online Exploration and Data Collection}

The agent continuously explores the environment and collects new experiences to populate the replay buffer $\mathcal{B}$.
Each episode is generated by first sampling an instruction $I_j$ from the self-generated instruction set and retrieving a corresponding goal image $g_j$ via the hybrid retrieval mechanism.
The agent then executes the current policy $\pi_\theta$ to generate a trajectory $\{(s_t, a_t, r_t, s_{t+1})\}_{t=0}^{T}$.
Finally, the trajectory is stored in the replay buffer $\mathcal{B}$, and the memory database $\mathcal{M}$ is updated with any newly observed keyframes.

This closed loop ensures that the policy is continuously refined based on the most recent environmental knowledge encoded in $\mathcal{M}$.

%% file: sections/5_experiment.tex
\section{Experiments and Analysis}

This section presents a comprehensive evaluation of \ours, our lifelong self-learning embodied navigation framework.
We demonstrate that \ours enables robots to achieve ``out-of-the-box'' navigation capabilities in completely unseen dynamic environments through autonomous exploration and continuous self-improvement, without any supervision or human intervention.

\subsection{Experimental Setup}

\subsubsection{Simulation and Real-World Platforms}

We conduct our primary simulation experiments using the Habitat simulator~\cite{savva2019habitat}.
The pre-training dataset consists of the training splits from HM3D~\cite{ramakrishnan2021habitat} and MP3D~\cite{chang2017matterport3d}.
For the main experimental evaluation, we select five distinct, unseen indoor scenes from these datasets, ensuring diverse layouts and spatial configurations to thoroughly assess generalization capabilities.
To validate the practical feasibility of our system, we also deploy and test \ours using a Unitree Go2 quadruped robot in three different real-world environments (laboratory, living room, and home).
These deployments demonstrate the system's ability to operate autonomously from initialization through exploration to responding to user instructions in realistic settings.

\subsubsection{Baseline Methods and Comparisons}

To provide a comprehensive evaluation, we compare \ours against the following state-of-the-art methods:

\textbf{ReMEmbR~\cite{anwar2025remembr}:} A Retrieval-augmented Memory for Embodied Robots, designed for lifelong video question answering for robot navigation.
ReMEmbR employs a structured approach involving a memory-building phase and an instruction-handling phase, leveraging temporal information, spatial information, and images to efficiently handle continuously growing robot histories.
To validate the effectiveness of our memory database, we replace the memory database component in \ours with ReMEmbR while keeping the RL component intact.

\textbf{ConRFT~\cite{chen2025conrft}:} A reinforced fine-tuning method for Vision-Language-Action (VLA) models via consistency policy.
ConRFT integrates behavior cloning and Q-learning in the offline stage, and employs consistency policy for online fine-tuning with human interventions.
To validate the effectiveness of our online RL method, we replace the VLA RL component in \ours with ConRFT while keeping the memory mechanisms intact.
Note that we test ConRFT without human interventions to ensure a fair comparison with \ours's fully autonomous operation.

\textbf{SERL~\cite{luo2024serl}:} A software suite for sample-efficient robotic reinforcement learning, providing a carefully implemented library containing a sample-efficient off-policy deep RL method.
SERL includes methods for computing rewards and resetting the environment, and has demonstrated efficient learning on various manipulation tasks.
Similar to ConRFT, we replace the online VLA RL component in \ours with SERL while keeping the memory mechanisms intact to validate our RL approach.

\textbf{OSG Navigator~\cite{loo2025open}:} A modular system for open-world Object-Goal Navigation that introduces the Open Scene Graph representation as spatial memory.
OSG Navigator organizes spatial information hierarchically using OSG schemas, which are templates describing the common structure of environment classes and can be automatically generated from simple semantic labels.
We fully reproduce OSG Navigator as a baseline with explicitly structured scene-graph memory, providing a contrast against our image-centric multimodal memory that does not encode explicit spatial graphs.

\subsubsection{Evaluation Metrics and Protocol}

\paragraph{Primary Metrics}
We employ three primary metrics to assess performance.
The \textbf{Success Rate (SR)} measures the percentage of episodes in which the agent autonomously stops within 300 steps and receives a positive reward from the reward model.
To evaluate navigation efficiency in simulation, we report \textbf{Success weighted by Path Length (SPL)}, which weights success by the ratio of the shortest path to the actual path length.
Additionally, we track the \textbf{Episode Length}, defined as the number of steps taken per episode, which reflects navigation efficiency in both simulation and real-world experiments.

\paragraph{Evaluation Protocol}
To ensure fair and rigorous evaluation, we adopt a strict protocol.
For each test scene, we predefine a fixed set of 100 open-vocabulary instructions that are never used during training.
During the training phase, we pause at regular intervals based on a fixed wall-clock time budget to evaluate model performance.
At each evaluation checkpoint, we run 100 test episodes from a fixed set of starting positions using the held-out instruction set.

\subsection{Implementation Details}
We specify the model architecture and system configurations used in our experiments.
Additional hyperparameters are provided in the appendix.

\subsubsection{Model Architecture}
The VLA backbone uses a Qwen2-7B transformer with 32 layers and a hidden dimension of 4096.
Visual encoding employs SigLIP and DINOv2.
A 2-layer MLP projects concatenated features to the model dimension.
The action head is a 4-layer MLP outputting 24 waypoint coordinates, and the critic is a 4-layer MLP.
Qwen3-VL-8B-Instruct is used for memory operations, and a frozen LLaVE-7B is employed for reward computation and embedding extraction.

\subsubsection{Computational Resources}
Pretraining was conducted on 40$\times$ NVIDIA H100 GPUs.
For online learning, we utilize 8$\times$ NVIDIA H100 GPUs.
The system runs at $\sim$5\,Hz on a single NVIDIA H100 GPU during real-world deployment.

\subsection{Main Results and Analysis}

\subsubsection{Lifelong Learning Performance}

\begin{figure*}[!t]
\centering
\includegraphics[width=0.95\textwidth]{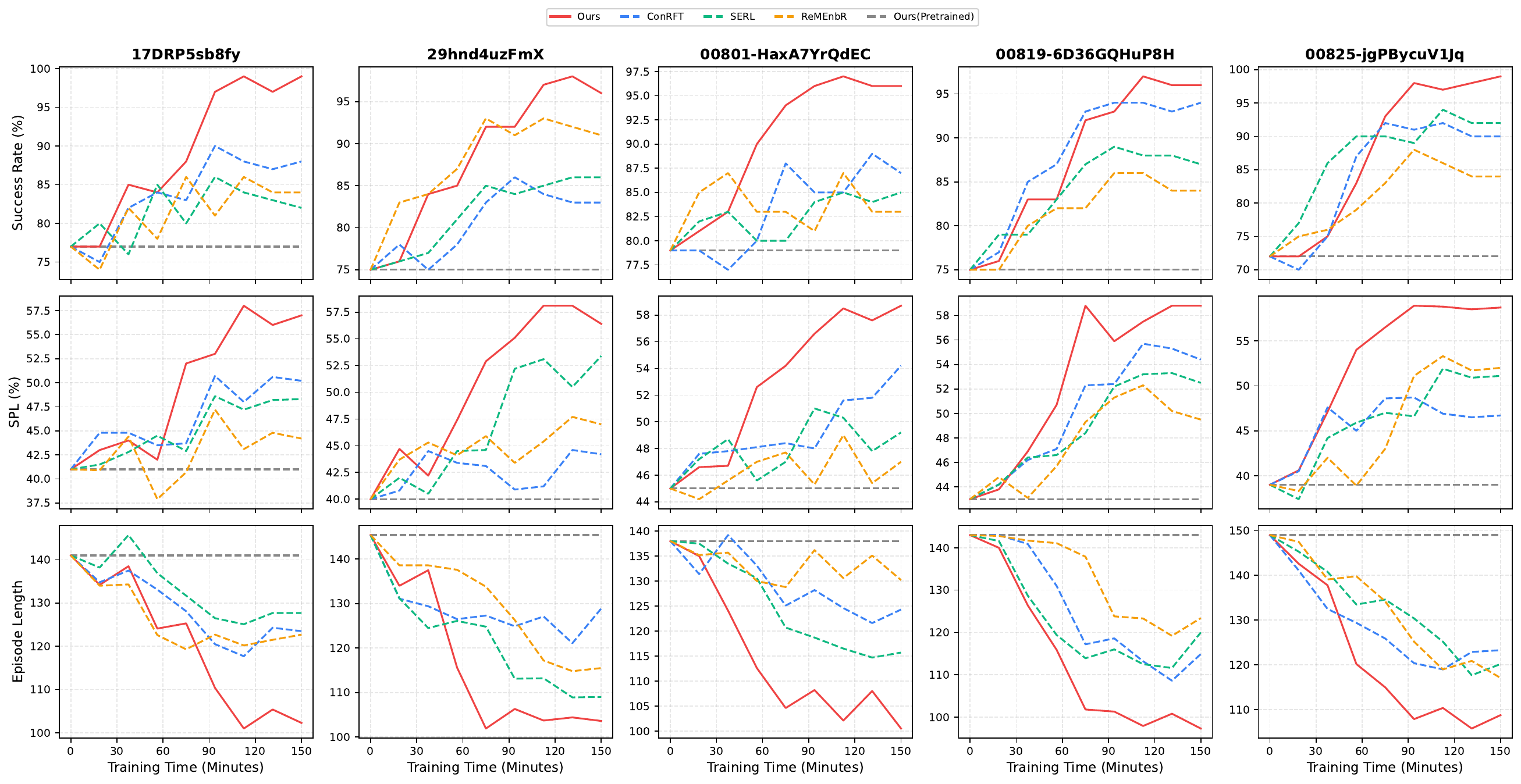}
\caption{Lifelong learning performance in simulation across five HM3D/MP3D test scenes. Each subplot reports the fixed-test performance of the pretrained model and subsequent online checkpoints sampled at equal wall-clock intervals for \ours and all learning-based baselines (ReMEmbR, ConRFT, SERL).}
\label{fig:lifelong_learning}
\end{figure*}

The core experiment demonstrates \ours's ability to progressively improve from its pretrained initialization to task mastery through autonomous exploration and self-learning.
Figure~\ref{fig:lifelong_learning} summarizes the lifelong learning trends across all HM3D/MP3D test scenes in simulation.

Figure~\ref{fig:lifelong_learning} reveals that \ours exhibits a steady upward trajectory across all three metrics, starting from performance near the pretrained model baseline and gradually converging to near-perfect success rates (approaching 100\%).
In contrast, baseline methods such as ReMEmbR, ConRFT, and SERL show limited improvement and plateau at significantly lower performance levels.
Because OSG Navigator is training-free, we report its fixed performance only in the summary tables.
The steady increase in SPL, together with the consistent decrease in episode length, indicates that \ours learns to navigate more efficiently over time, taking shorter and more direct paths to goals with higher path efficiency.
This performance gain is a direct consequence of the accumulated experience in the memory database and the improved policy learned through online updates.
The results demonstrate that \ours's continuous learning mechanism enables it to adapt and excel in unseen environments, significantly outperforming both learning-based and static baseline methods.

Table~\ref{tab:final_performance} presents the final quantitative performance comparison across all methods on the standard test set after training convergence.

\begin{table*}[!t]
\renewcommand{\arraystretch}{1.15}
\caption{Final Performance Comparison in Simulation Environments. Columns are grouped by five unseen scenes (S1--S5); each scene reports SR, SPL, and Episode Length (Len) after convergence.}
\label{tab:final_performance}
\centering
\begin{tabular}{l|ccc|ccc|ccc|ccc|ccc}
\hline
& \multicolumn{3}{c|}{\textbf{Scene S1}} & \multicolumn{3}{c|}{\textbf{Scene S2}} & \multicolumn{3}{c|}{\textbf{Scene S3}} & \multicolumn{3}{c|}{\textbf{Scene S4}} & \multicolumn{3}{c}{\textbf{Scene S5}} \\
\textbf{Method} & \textbf{SR} & \textbf{SPL} & \textbf{Len} & \textbf{SR} & \textbf{SPL} & \textbf{Len} & \textbf{SR} & \textbf{SPL} & \textbf{Len} & \textbf{SR} & \textbf{SPL} & \textbf{Len} & \textbf{SR} & \textbf{SPL} & \textbf{Len} \\
\hline
ReMEmbR~\cite{anwar2025remembr} & 84.0 & 44.2 & 122.7 & 91.0 & 47.0 & 115.5 & 83.0 & 47.0 & 130.2 & 84.0 & 49.5 & 123.4 & 84.0 & 52.0 & 117.1 \\
ConRFT~\cite{chen2025conrft} & 88.0 & 50.2 & 123.5 & 83.0 & 44.2 & 128.9 & 87.0 & 54.2 & 124.3 & 94.0 & 54.4 & 114.9 & 90.0 & 46.7 & 123.3 \\
SERL~\cite{luo2024serl} & 82.0 & 48.3 & 127.7 & 86.0 & 53.4 & 109.0 & 85.0 & 49.2 & 115.7 & 87.0 & 52.5 & 120.0 & 92.0 & 51.1 & 120.2 \\
OSG Navigator~\cite{loo2025open} & 75.0 & 39.2 & 134.7 & 76.0 & 39.2 & 135.9 & 76.0 & 42.0 & 137.2 & 77.0 & 44.5 & 130.4 & 77.0 & 41.7 & 130.3 \\
\textbf{\ours (Ours)} & \textbf{99.0} & \textbf{57.0} & \textbf{102.3} & \textbf{96.0} & \textbf{56.4} & \textbf{103.6} & \textbf{96.0} & \textbf{58.7} & \textbf{100.5} & \textbf{96.0} & \textbf{58.8} & \textbf{97.3} & \textbf{99.0} & \textbf{58.7} & \textbf{108.8} \\
\hline
\end{tabular}
\end{table*}

\ours achieves the highest success rate and SPL, along with the shortest episode length, establishing clear superiority over all baseline methods.
Notably, \ours significantly outperforms both learning-based baselines (ReMEmbR, ConRFT, SERL) and the static memory-based method (OSG Navigator), highlighting the advantage of continuous learning over both static map-based approaches and alternative learning strategies.

\begin{figure*}[!t]
\centering
\includegraphics[width=0.95\textwidth]{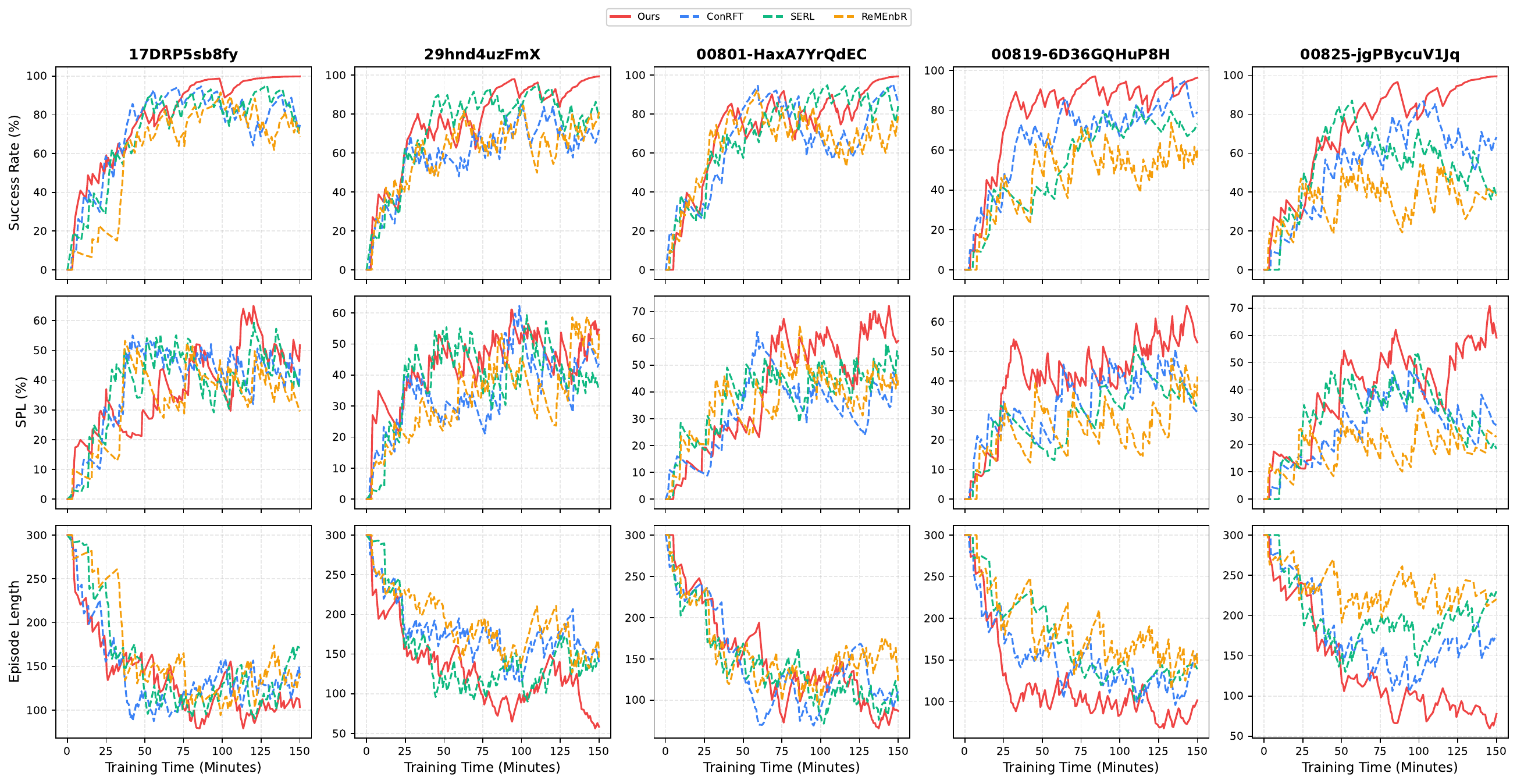}
\caption{Online exploration performance during autonomous learning. This figure shows the dynamic performance on self-instruction tasks during exploration using Exponential Moving Average (EMA) of Success Rate (SR), SPL, and Episode Length.}
\label{fig:online_success}
\end{figure*}

To further demonstrate the learning dynamics during autonomous exploration, Figure~\ref{fig:online_success} shows the online performance during exploration---the real-time performance on self-generated navigation tasks as the agent actively explores and learns in the environment.
Unlike Figure~\ref{fig:lifelong_learning}, which reports fixed test set averages, Figure~\ref{fig:online_success} plots Exponential Moving Average (EMA) metrics (SR, SPL, Episode Length) computed directly from the online exploration stream.
These metrics capture the real-time learning behavior during autonomous exploration.

\subsubsection{Ablation Studies}

\begin{figure*}[!t]
\centering
\includegraphics[width=0.95\textwidth]{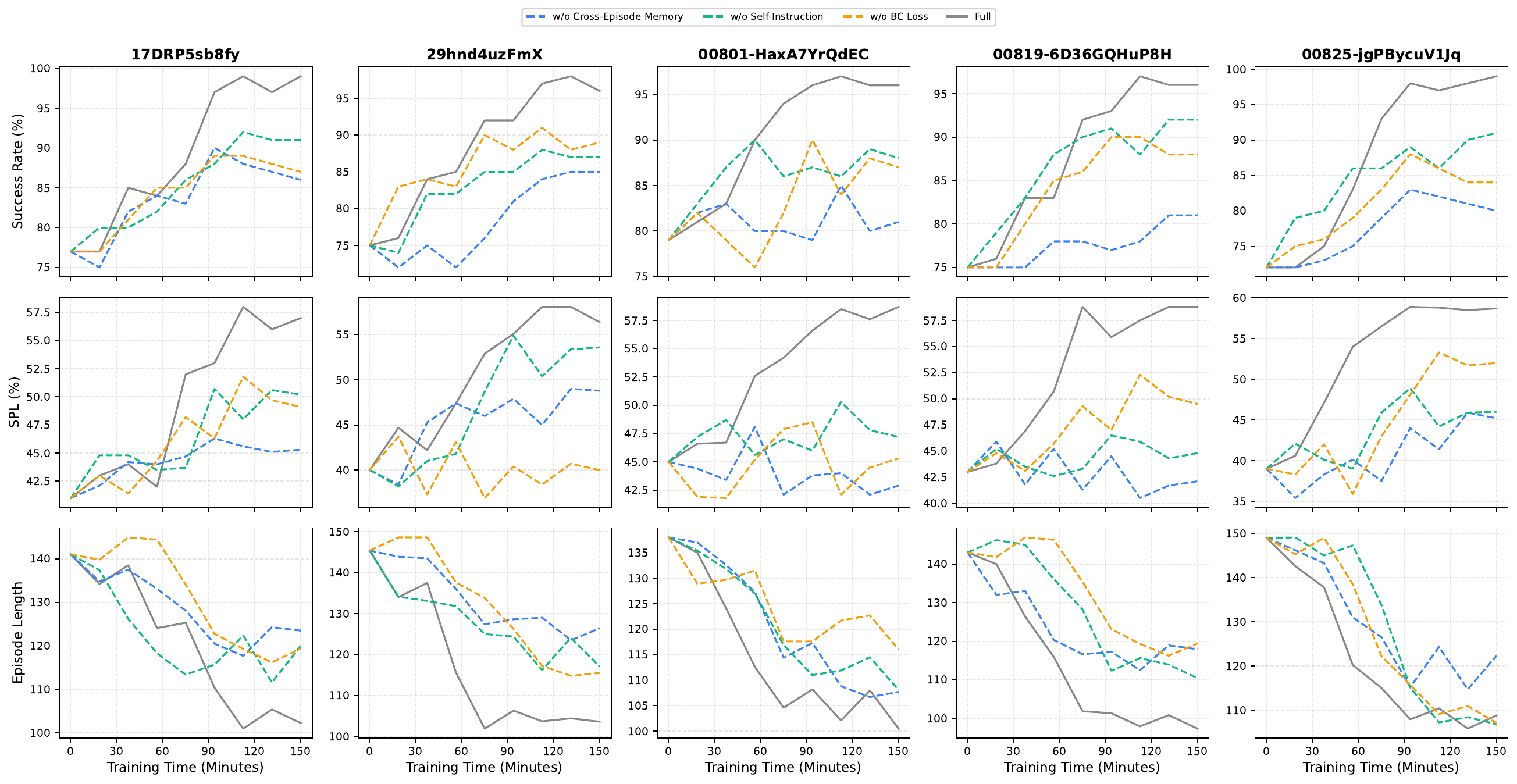}
\caption{Ablation study results comparing the full \ours model with ablation variants across three key metrics: Success Rate (SR), Success weighted by Path Length (SPL), and Episode Length.}
\label{fig:ablation}
\end{figure*}

To validate the contribution of each core component, we conduct comprehensive ablation studies that isolate the key mechanisms of \ours. We evaluate three variants, including two memory-component ablations and one RL-component ablation:

\textbf{w/o Cross-Episode Memory:} The memory buffer is cleared after every episode, preventing knowledge accumulation across long horizons. This variant shows markedly slower improvement and lower final performance (Figure~\ref{fig:ablation}), confirming that retaining experience across episodes is critical for lifelong learning.

\textbf{w/o Self-Instruction:} The self-instruction generator is replaced with random goal sampling, which strips the agent of semantically meaningful training signals. As a result, success rates and SPL remain substantially lower (Figure~\ref{fig:ablation}), demonstrating that self-instruction is key to producing diverse, curriculum-like tasks.

\textbf{w/o BC Loss:} We remove the behavior cloning (BC) loss during online fine-tuning, relying solely on the reinforcement learning objective. This variant exhibits unstable learning dynamics and slower convergence (Figure~\ref{fig:ablation}), indicating that the BC loss acts as a crucial regularizer that stabilizes updates and prevents catastrophic forgetting of the pre-trained prior.

Figure~\ref{fig:ablation} visualizes the aggregated trends for these variants, highlighting the contribution of each component relative to the full model.

\begin{table*}[!t]
\renewcommand{\arraystretch}{1.15}
\caption{Quantitative Results of Ablation Studies across Five Scenes (S1--S5). Each scene column contains SR, SPL, and Episode Length (Len).}
\label{tab:ablation}
\centering
\begin{tabular}{l|ccc|ccc|ccc|ccc|ccc}
\hline
& \multicolumn{3}{c|}{\textbf{Scene S1}} & \multicolumn{3}{c|}{\textbf{Scene S2}} & \multicolumn{3}{c|}{\textbf{Scene S3}} & \multicolumn{3}{c|}{\textbf{Scene S4}} & \multicolumn{3}{c}{\textbf{Scene S5}} \\
\textbf{Model Variant} & \textbf{SR} & \textbf{SPL} & \textbf{Len} & \textbf{SR} & \textbf{SPL} & \textbf{Len} & \textbf{SR} & \textbf{SPL} & \textbf{Len} & \textbf{SR} & \textbf{SPL} & \textbf{Len} & \textbf{SR} & \textbf{SPL} & \textbf{Len} \\
\hline
w/o Cross-Episode Memory & 86.0 & 45.3 & 123.5 & 85.0 & 48.8 & 126.4 & 81.0 & 42.9 & 107.7 & 81.0 & 42.1 & 117.9 & 80.0 & 45.2 & 122.3 \\
w/o Self-Instruction & 91.0 & 50.2 & 120.0 & 87.0 & 53.6 & 117.1 & 88.0 & 47.2 & 108.1 & 92.0 & 44.8 & 110.4 & 91.0 & 46.0 & 106.8 \\
w/o BC Loss & 87.0 & 49.1 & 119.4 & 89.0 & 40.0 & 115.5 & 87.0 & 45.3 & 116.0 & 88.0 & 49.5 & 119.4 & 84.0 & 52.0 & 107.1 \\
\textbf{\ours (Full)} & \textbf{99.0} & \textbf{57.0} & \textbf{102.3} & \textbf{96.0} & \textbf{56.4} & \textbf{103.6} & \textbf{96.0} & \textbf{58.7} & \textbf{100.5} & \textbf{96.0} & \textbf{58.8} & \textbf{97.3} & \textbf{99.0} & \textbf{58.7} & \textbf{108.8} \\
\hline
\end{tabular}
\end{table*}

Table~\ref{tab:ablation} provides precise quantitative results for the ablation study, supporting the observations from Figure~\ref{fig:ablation}.
The full \ours model consistently outperforms all ablation variants across all metrics, confirming that the integration of online updates, cross-episode memory, and self-instruction generation is necessary for achieving the best performance.

\subsection{Generalization and Robustness Analysis}

\subsubsection{Adaptability to Dynamic Environments}

A critical requirement for real-world deployment is the ability to adapt to environmental changes.
To evaluate this capability, we simulate dynamic environments by suddenly changing the scene after the agent has converged in the initial environment.
This abrupt change tests the agent's ability to adapt to environmental shifts.

\begin{figure*}[!t]
\centering
\includegraphics[width=0.95\textwidth]{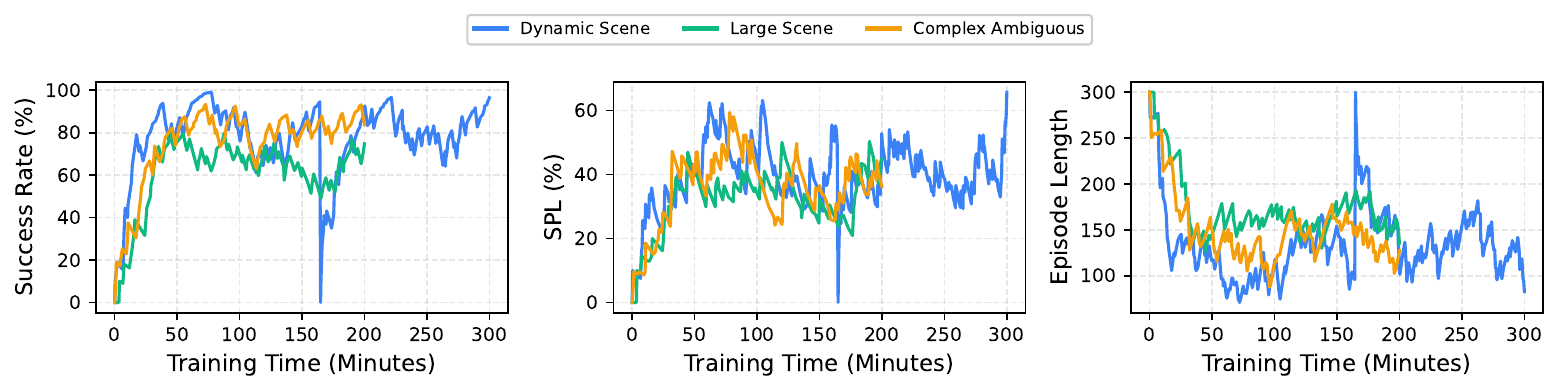}
\caption{Online learning performance on challenging scenarios. The figure shows the Exponential Moving Average (EMA) of Success Rate, SPL, and Episode Length during autonomous exploration for three challenging task categories: Dynamic Scene Navigation (left), Large Scene Navigation (middle), and Complex Ambiguous Instructions (right).}
\label{fig:complex_tasks_online}
\end{figure*}

Figure~\ref{fig:complex_tasks_online} (left) shows that the performance drops sharply immediately after the scene change, but then rapidly recovers and converges again through continued exploration.
This demonstrates the model's strong adaptability to dynamic environments.
The continuous online update mechanism enables \ours to quickly adapt to environmental changes by updating its memory database with new observations and refining its policy accordingly.
In contrast, the ablation variant without cross-episode memory (w/o Cross-Episode Memory) shows persistent performance degradation and fails to recover, highlighting the critical role of continuous learning in handling dynamic environments.
This experiment validates that \ours's online update mechanism enables the system to adapt to environmental changes in real-time, a crucial capability for practical deployment.

\subsubsection{Generalization to Challenging Scenarios}

To further assess \ours's generalization capabilities, we evaluate performance on two additional challenging task categories.
First, we validate the model on \textbf{Large Scene Navigation} using an ultra-large scene from the MP3D dataset.
The learning curve shows that convergence in large scenes is slower and requires longer training time compared to standard scenes.
Second, for \textbf{Complex Ambiguous Instructions}, the ambiguity leads to more random object retrieval, which to some extent affects training efficiency, resulting in slower convergence.

Figure~\ref{fig:complex_tasks_online} (middle and right) visualizes the online learning dynamics for large scene navigation and complex ambiguous instructions, showing how \ours progressively adapts and improves through autonomous exploration.
The curves demonstrate that \ours maintains stable learning trajectories even under these challenging conditions, with Success Rate steadily increasing and Episode Length decreasing over time.

\begin{table}[!t]
\renewcommand{\arraystretch}{1.15}
\caption{Online Performance Evolution on Challenging Scenarios. We report the Exponential Moving Average (EMA) of Success Rate (SR) and Episode Length (Len) during autonomous exploration.}
\label{tab:complex_tasks}
\centering
\resizebox{\columnwidth}{!}{
\begin{tabular}{l|cc|cc|cc}
\hline
& \multicolumn{2}{c|}{\textbf{Dynamic}} & \multicolumn{2}{c|}{\textbf{Large}} & \multicolumn{2}{c}{\textbf{Complex}} \\
& \multicolumn{2}{c|}{\textbf{Scene}} & \multicolumn{2}{c|}{\textbf{Scene}} & \multicolumn{2}{c}{\textbf{Ambiguous}} \\
\textbf{Learning Stage} & \textbf{SR} & \textbf{Len} & \textbf{SR} & \textbf{Len} & \textbf{SR} & \textbf{Len} \\
\hline
Early Stage & 71.2 & 125.1 & 69.5 & 145.3 & 72.8 & 149.3 \\
Mid Stage (Pre-change) & 98.9 & 83.2 & 73.6 & 137.2 & 82.3 & 145.4 \\
Scene Change / Late & 68.5 & 122.4 & - & - & - & - \\
\textbf{Final Stage} & \textbf{96.4} & \textbf{82.9} & \textbf{74.7} & \textbf{135.5} & \textbf{83.8} & \textbf{127.5} \\
\hline
\end{tabular}
}
\end{table}

Table~\ref{tab:complex_tasks} presents the final quantitative performance comparison after convergence, including all three challenging task categories (dynamic scene, large scene, and complex ambiguous instructions).
\ours demonstrates superior performance across all challenging task categories.
In large-scale scenes, the accumulated spatial knowledge in the memory database allows the agent to plan efficient multi-step paths and maintain coherent lifelong strategies.
For complex ambiguous instructions, the rich semantic understanding provided by the vision-language model, combined with diverse training signals from self-instruction generation, enables effective handling of semantically challenging instructions.

\subsection{Internal Mechanism Analysis}

\subsubsection{Exploration Visualization and Retrieval Analysis}

\begin{figure}[!t]
\centering
\includegraphics[width=0.95\columnwidth]{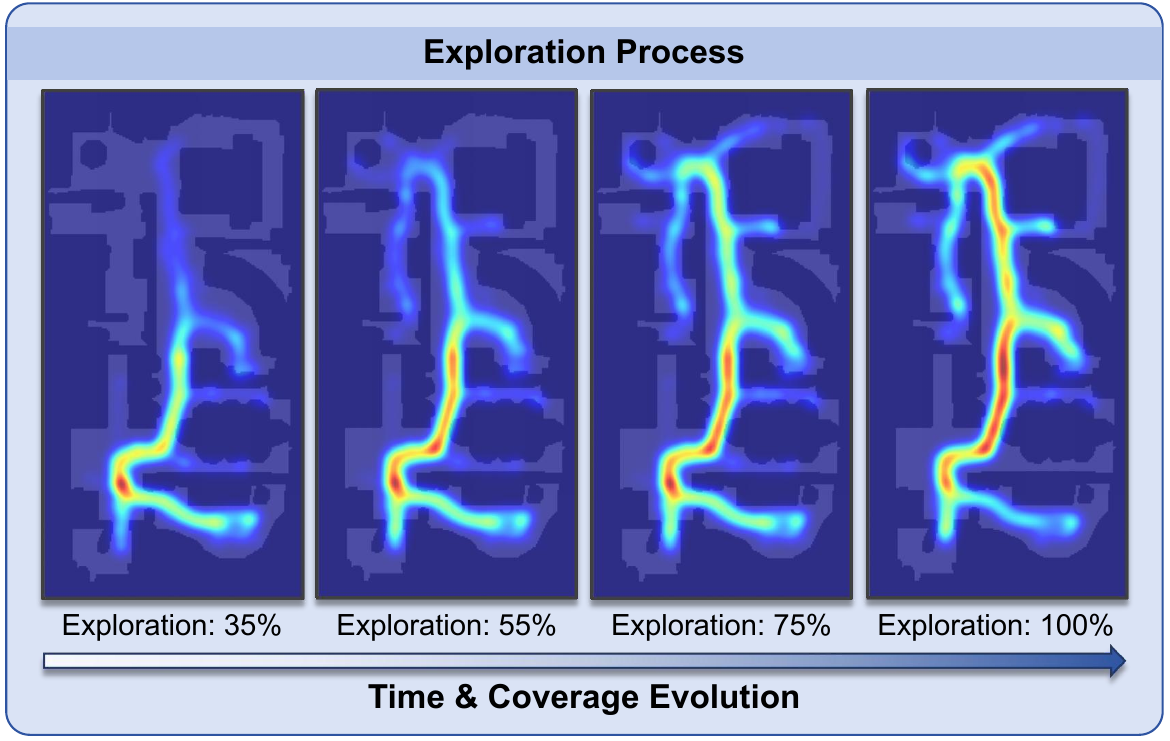}
\caption{Visualization of the exploration process in Scene S1 using trajectory heatmaps. The four subplots correspond to the initial, early, late, and final stages of exploration.}
\label{fig:heatmap}
\end{figure}

To visualize the autonomous exploration process, Figure~\ref{fig:heatmap} presents the exploration heatmaps in Scene S1 across four distinct stages: initial, early, late, and final.
These heatmaps depict the frequency of the robot's presence at different locations on the top-down map, illustrating the progression of spatial coverage over time.

As shown in Figure~\ref{fig:heatmap}, the exploration begins with coverage confined to a local area around the starting point.
During the early and late stages, the robot progressively pushes its exploration boundaries to visit previously unseen regions.
By the final stage, the heatmap indicates comprehensive coverage of the entire environment.
The heatmaps demonstrate how the robot's exploration coverage expands from a local region to the entire environment, validating the effectiveness of the autonomous exploration strategy.
This gradual expansion confirms that \ours successfully transitions from local exploration to global scene understanding, demonstrating the system's capability to autonomously explore and map complex indoor environments without human supervision.

\begin{figure}[!t]
\centering
\includegraphics[width=0.95\columnwidth]{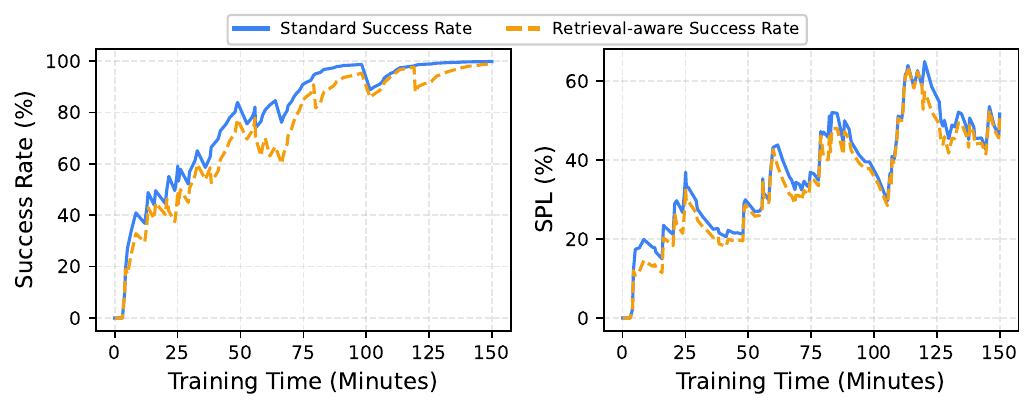}
\caption{Retrieval accuracy analysis during online exploration in Scene S1. The figure compares the standard online success rate (solid line) with the retrieval-aware success rate (dashed line) where success requires both accurate retrieval (human-verified) and goal achievement.}
\label{fig:retrieval_accuracy}
\end{figure}

\textbf{Quantitative Retrieval Accuracy Analysis:}
To quantitatively assess the retrieval quality, we conduct a human evaluation study where annotators manually judge whether the retrieved goal images correctly match the given instructions.
Figure~\ref{fig:retrieval_accuracy} compares two online success rate curves during autonomous exploration: (1) the standard success rate that considers an episode successful if the agent reaches the retrieved goal image, and (2) the retrieval-aware success rate that requires both accurate retrieval (human-verified) and successful navigation to count as success.

The results show that the retrieval-aware success rate curve is only slightly lower than the standard success rate curve, with a consistent small gap throughout the exploration process.
The small gap between the two curves demonstrates that the multimodal memory retrieval mechanism achieves high accuracy, though not perfect, validating the effectiveness of the vision-language understanding and semantic matching components.
This small gap indicates that the vast majority of retrievals are semantically correct, demonstrating the high accuracy of the multimodal memory retrieval mechanism.
The fact that the gap is non-zero reflects realistic limitations in vision-language understanding for highly ambiguous instructions, but the overall high alignment confirms that retrieval errors are not a primary bottleneck in the system's performance.
This analysis validates that \ours's retrieval component effectively bridges natural language instructions to visual goal representations, supporting the end-to-end navigation performance reported in previous sections.

\begin{figure}[!t]
\centering
\includegraphics[width=0.95\columnwidth]{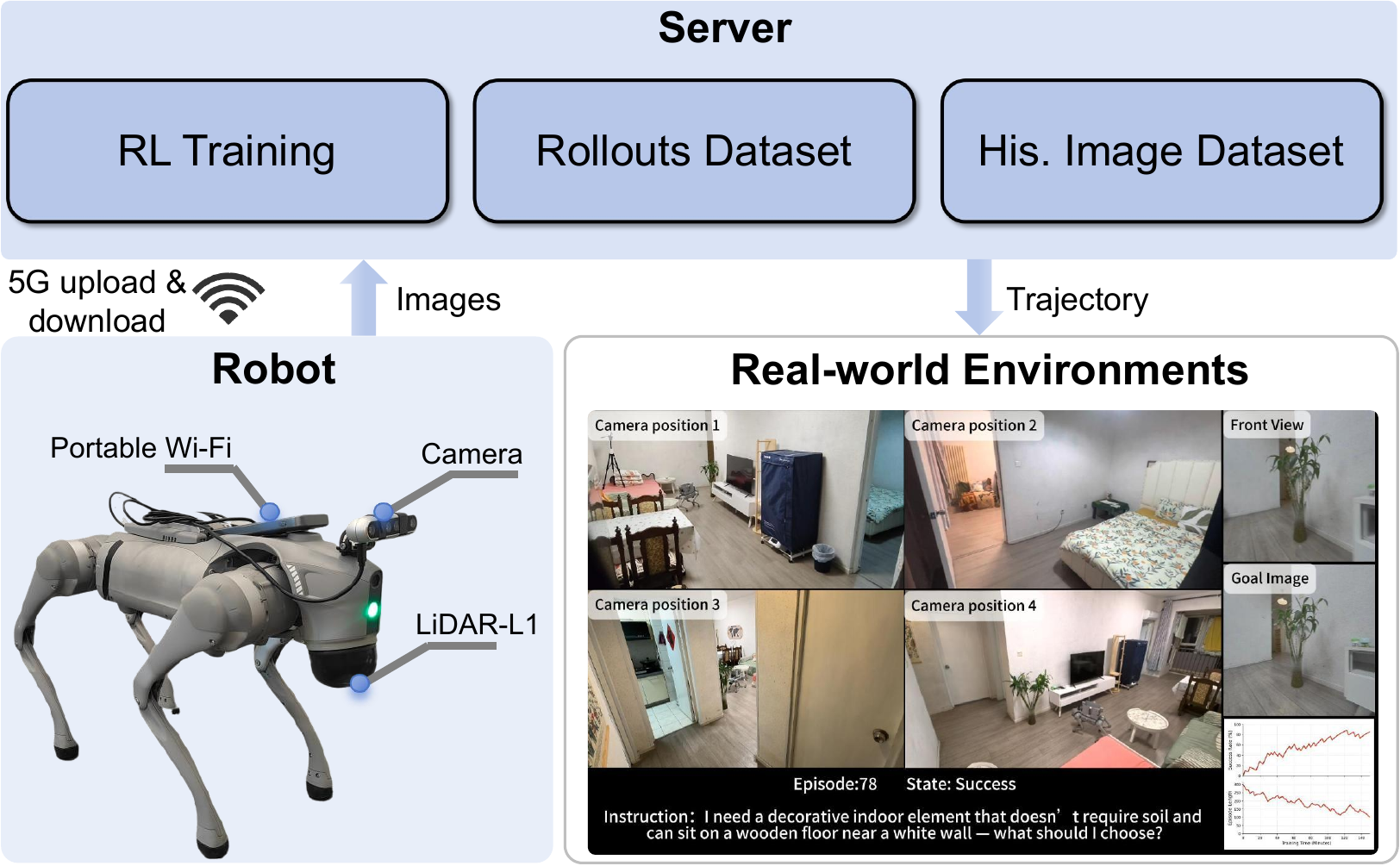}
\caption{Real-world experimental setup. The Unitree Go2 platform is augmented with a high-FOV RGB camera, an LiDAR-L1 scanner for obstacle avoidance, and a 5G communication module. The deployment architecture shows that sensor data (RGB, LiDAR) is streamed via the 5G module to the remote H100 server running \ours, while a web server relays control commands back to the real-world environments.}
\label{fig:real_robot}
\end{figure}

\subsubsection{Memory Growth Analysis}

To demonstrate the knowledge accumulation process, Table~\ref{tab:memory_growth} reports the growth of the memory database size over different exploration stages in Scene S1.

\begin{table}[!t]
\renewcommand{\arraystretch}{1.15}
\caption{Memory database size evolution over exploration stages in Scene S1.}
\label{tab:memory_growth}
\centering
\begin{tabular}{l|ccccc}
\hline
\textbf{Stage} & \textbf{Initial} & \textbf{Early} & \textbf{Mid} & \textbf{Late} & \textbf{Final} \\
\hline
\textbf{Memory Size} & 0 & 23 & 39 & 42 & 44 \\
\hline
\end{tabular}
\end{table}

As the memory database grows through autonomous exploration, the agent gains access to a richer set of semantic locations and spatial relationships.
This continuous accumulation of knowledge enables more accurate instruction understanding and target retrieval, serving as the foundation for the system's lifelong learning capability.

\begin{figure*}[!t]
\centering
\includegraphics[width=0.95\textwidth]{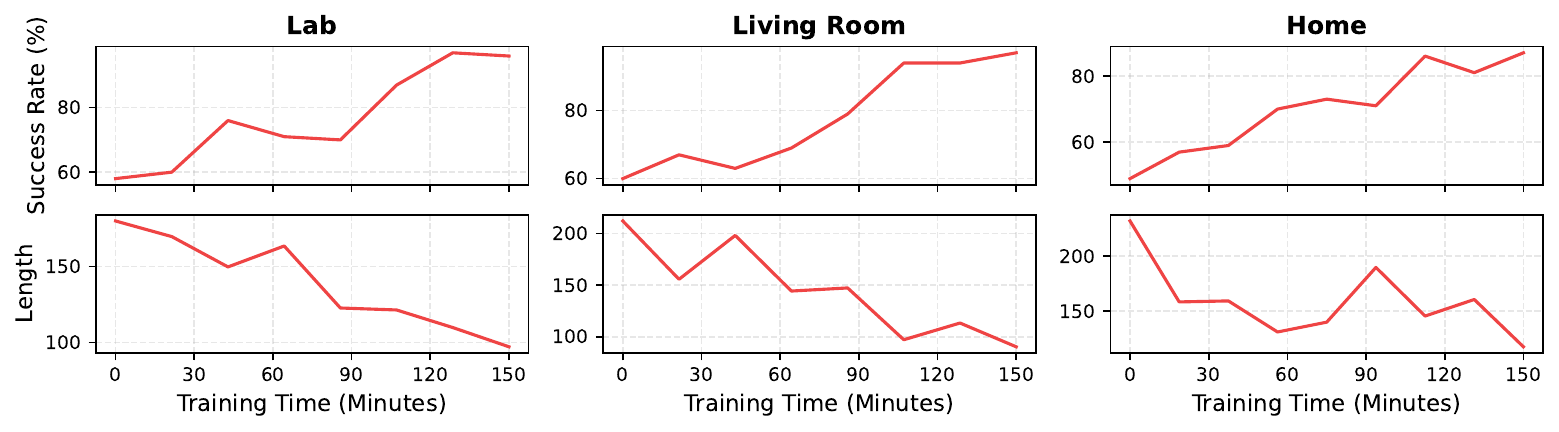}
\caption{Real-world lifelong learning performance across three real-world environments.
Rows correspond to Success Rate and Episode Length; columns correspond to the three environments.
Each subplot shows the performance evolution of \ours from initialization to convergence.}
\label{fig:realworld_learning}
\end{figure*}

\begin{figure*}[!t]
\centering
\includegraphics[width=0.95\textwidth]{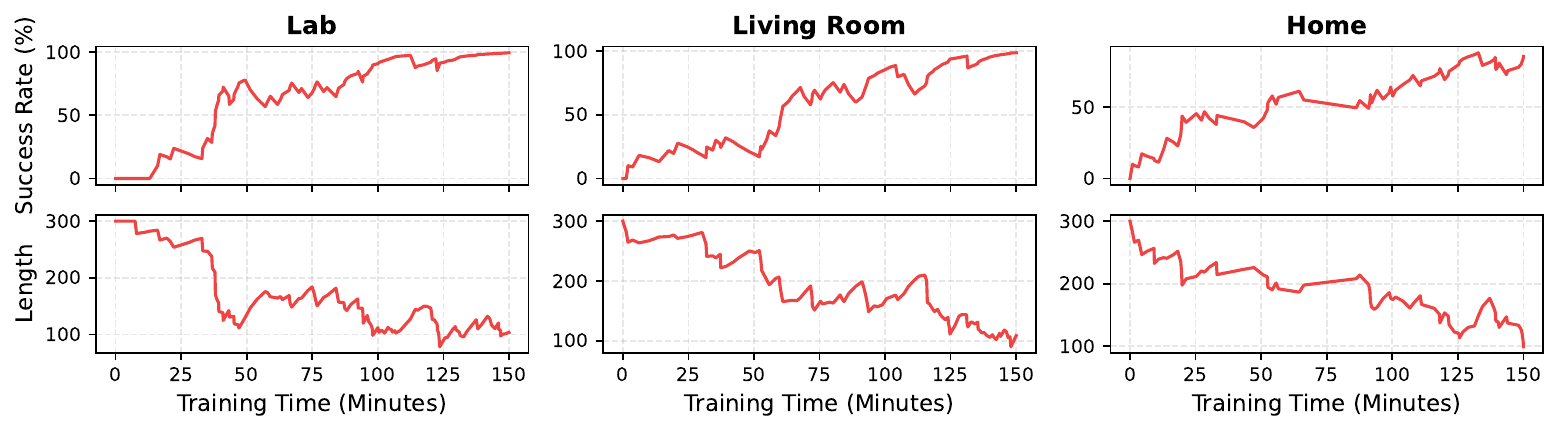}
\caption{Real-world online exploration performance using EMA of Success Rate and Episode Length during autonomous deployment.
These metrics capture real-time learning behavior in each physical environment.}
\label{fig:realworld_online}
\end{figure*}

\subsection{Real-World Experiments}

We deploy \ours in three real-world environments (laboratory, living room, and home) using a Unitree Go2 quadruped.
We detail the concrete hardware and system integration that enable fully autonomous operation.

\textbf{Robot Hardware and System Architecture:}
Figure~\ref{fig:real_robot} illustrates the experimental setup, comprising the hardware platform and the deployment architecture.

We attach a forward-facing RGB camera (height 0.6\,m, 120° FOV) whose images are undistorted before being fed to the VLA policy, a 5G communication module that streams data to the remote server with low latency, and an LiDAR-L1 unit that supplies dense depth signals for obstacle avoidance and safe navigation.
These components collectively allow the robot to perceive cluttered indoor spaces while maintaining reliable connectivity during long deployments.

Regarding the system architecture, sensor streams are transmitted through the 5G module to a remote H100 server that hosts \ours's memory and policy learning stack.
A lightweight web server bridges the robot and the remote compute, forwarding high-level actions back to the Unitree controller while logging trajectories in the real-world environment.
This architecture mirrors the simulation pipeline, enabling consistent lifelong learning behaviors in physical spaces without any additional model adaptation.
The deployment demonstrates the complete autonomous pipeline: from system initialization, through autonomous exploration and memory database construction, to responding to user instructions.

\textbf{Real-World Learning and Quantitative Performance:}
Figure~\ref{fig:realworld_learning} reports the real-world lifelong learning curves, tracking the agent's progress from initialization, while Table~\ref{tab:realworld} summarizes the performance evolution across all environments.

We visualize Success Rate and Episode Length across three physical environments using the same layout as the simulation plots.
\ours starts close to the pretrained performance level and steadily improves toward high success rates while reducing episode length, demonstrating that the self-learning pipeline remains effective despite network latency, control inaccuracies, and other sim-to-real gaps.

\textbf{Online Exploration Performance:}
Figure~\ref{fig:realworld_online} shows the Exponential Moving Average (EMA) of Success Rate and Episode Length during real-world exploration.

These curves highlight \ours's ability to maintain stable performance while continuously updating its policy on-device.
The EMA smoothing emphasizes the trend that \ours improves over time, successfully adapting to the physical environment through continuous learning.

\begin{table*}[!t]
\renewcommand{\arraystretch}{1.15}
\caption{Real-World Navigation Performance Evolution across Three Environments. We report both fixed-set evaluation metrics (Eval) and online exploration metrics (Online EMA) across the Laboratory (R1), Living Room (R2), and Home (R3). The table tracks the performance improvement of \ours from initialization to convergence.}
\label{tab:realworld}
\centering
\setlength{\tabcolsep}{8.5pt}
\begin{tabular}{l|cccc|cccc|cccc}
\hline
& \multicolumn{4}{c|}{\textbf{Lab (R1)}} & \multicolumn{4}{c|}{\textbf{Living Room (R2)}} & \multicolumn{4}{c}{\textbf{Home (R3)}} \\
& \multicolumn{2}{c}{\textbf{Eval}} & \multicolumn{2}{c|}{\textbf{Online}} & \multicolumn{2}{c}{\textbf{Eval}} & \multicolumn{2}{c|}{\textbf{Online}} & \multicolumn{2}{c}{\textbf{Eval}} & \multicolumn{2}{c}{\textbf{Online}} \\
\textbf{Learning Stage} & \textbf{SR} & \textbf{Len} & \textbf{SR} & \textbf{Len} & \textbf{SR} & \textbf{Len} & \textbf{SR} & \textbf{Len} & \textbf{SR} & \textbf{Len} & \textbf{SR} & \textbf{Len} \\
\hline
Initialization & 58.0 & 180.1 & 0.0 & 300.0 & 60.0 & 212.2 & 0.0 & 300.0 & 49.0 & 231.6 & 0.0 & 300.0 \\
Early Stage & 76.0 & 149.8 & 23.9 & 253.9 & 63.0 & 198.2 & 24.9 & 274.1 & 59.0 & 159.3 & 30.5 & 235.3 \\
Mid Stage & 70.0 & 122.7 & 64.1 & 177.8 & 79.0 & 147.3 & 72.6 & 164.4 & 73.0 & 140.1 & 54.6 & 213.9 \\
\textbf{Final Stage} & \textbf{96.0} & \textbf{97.1} & \textbf{99.1} & \textbf{108.0} & \textbf{97.0} & \textbf{90.3} & \textbf{98.7} & \textbf{108.4} & \textbf{87.0} & \textbf{117.8} & \textbf{85.6} & \textbf{99.2} \\
\hline
\end{tabular}
\end{table*}

Table~\ref{tab:realworld} further shows that \ours achieves consistently high Success Rate across all environments when evaluated on the fixed set of 100 held-out instructions, indicating robust generalization in the physical world.

\begin{figure*}[!t]
\centering
\includegraphics[width=0.95\textwidth]{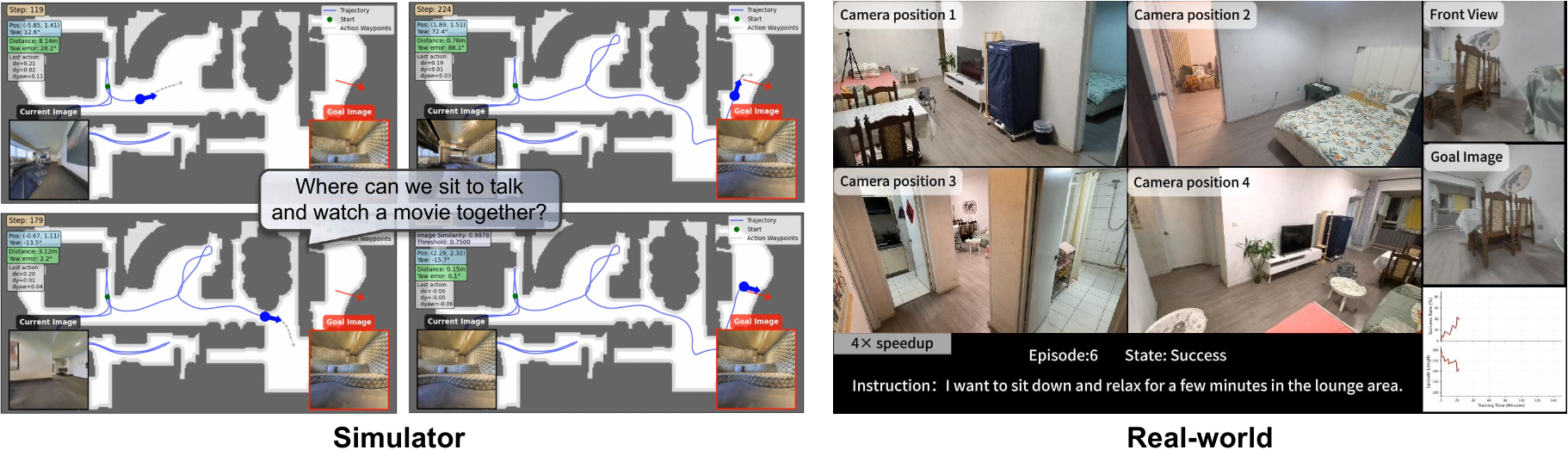}
\caption{Visualization of navigation episodes. The figure displays a navigation trajectory in the simulation environment (left) and a real-world deployment case (right).}
\label{fig:visualization}
\end{figure*}

\textbf{Qualitative Results:}
Figure~\ref{fig:visualization} visualizes successful navigation cases, showing both a simulation trajectory and a real-world experiment.
These demonstrations illustrate the robot's ability to: (1) autonomously explore and build memory of new environments, (2) understand natural language instructions from users, (3) retrieve relevant target locations from its memory, and (4) execute successful navigation trajectories.
The figure demonstrates the system's capability to operate effectively in both simulated and physical domains.
The qualitative results confirm that \ours operates effectively in real-world settings with realistic sensor noise, lighting variations, and dynamic obstacles.

The real-world experiments validate that \ours's design principles translate effectively from simulation to physical deployment, confirming the system's potential for practical applications in real-world service robotics scenarios.

%% file: sections/6_conclusion.tex
\section{Conclusion}

This paper introduced \ours, a lifelong self-learning embodied navigation framework that autonomously acquires open-vocabulary navigation skills in initially unknown and dynamically changing environments without human supervision.
By addressing two key limitations of existing vision-language-action models—the lack of persistent multimodal memory and the inability to adapt policies continuously through interaction—\ours enables embodied agents to construct stable internal representations from fragmented egocentric observations and to improve their navigation behavior over time.

\ours combines a pretrained VLA backbone with dual visual encoders and history compression, a self-evolving multimodal memory database with diversity-preserving updates, a self-instruction mechanism that turns accumulated scene knowledge into rich curricula via hybrid memory–internet goal retrieval, and a conservative Q-learning module that enables stable policy optimization from mixed online–offline data.
Extensive experiments in simulated and real-world environments show that \ours consistently outperforms strong baselines in terms of success rate, path efficiency, and generalization to complex open-vocabulary instructions, while maintaining robust performance under environmental changes.
Deployment on a quadruped robot further demonstrates that the proposed sim-to-real strategy allows the learned policy to transfer effectively to physical platforms with realistic perception and control noise, yielding reliable long-horizon navigation in the wild.

Our ablation studies verify that memory persistence, self-instruction, and conservative online updates are all necessary for achieving sustained performance gains, and that their integration produces a synergistic effect that underpins truly autonomous lifelong learning.
Nevertheless, \ours still relies on powerful vision-language models and substantial computation, and its exploration behavior is not explicitly constrained for safety-critical settings.
Future work will explore more efficient and safety-aware exploration strategies, tighter integration with human feedback, and extensions to multi-agent and multi-task scenarios, pushing embodied agents closer to deployment in complex, open-world service environments.